\documentclass[review]{elsarticle}

\usepackage{lineno,hyperref}

\usepackage[justification=centering]{caption}
\usepackage{subfig}

\modulolinenumbers[5]

\journal{Journal of \LaTeX\ Templates}

\begin{document}

\begin{frontmatter}

\title{Review of the Fingerprint Liveness Detection (LivDet) Competition Series: 2009 to 2015}


%%% Group authors per affiliation:


\author[mymainaddress]{Luca Ghiani}
\ead{luca.ghiani@diee.unica.it}

\author[mysecondaryaddress]{David A. Yambay}
\ead{yambayda@clarkson.edu}

\author[mymainaddress]{Valerio Mura}
\ead{valerio.mura@diee.unica.it}

\author[mymainaddress]{Gian Luca Marcialis}
\ead{marcialis@diee.unica.it}

\author[mymainaddress]{Fabio Roli}
\ead{roli@diee.unica.it}

\author[mysecondaryaddress]{Stephanie A. Schuckers}
\ead{sschucke@clarkson.edu}

\address[mymainaddress]{University of Cagliari (Italy)\\
Department of Electrical and Electronic Engineering}
\address[mysecondaryaddress]{Clarkson University (USA)\\
Department of Electrical and Computer Engineering}

\begin{abstract}
A spoof attack, a subset of presentation attacks, is the use of an artificial replica of a biometric in an attempt to circumvent a biometric sensor.
Liveness detection, or presentation attack detection, distinguishes between live and fake biometric traits and is based on the principle that additional information can be garnered above and beyond the data procured by a standard authentication system to determine if a biometric measure is authentic.

%The Fingerprint Liveness Detection Competition (LivDet) goal is to compare both software-based and hardware-based fingerprint liveness detection methodologies, using a standardized testing protocol and large quantities of spoof and live images. The competition is open to all academic and industrial institutions which have a solution for either software-based or system-based vitality detection problem.
The goals for the Liveness Detection (LivDet) competitions are to compare software-based fingerprint liveness detection and artifact detection algorithms (Part 1), as well as fingerprint systems which incorporate liveness detection or artifact detection capabilities (Part 2), using a standardized testing protocol and large quantities of spoof and live tests. The competitions are open to all academic and industrial institutions which have a solution for either software-based or system-based fingerprint liveness detection. The LivDet competitions have been hosted in 2009, 2011, 2013 and 2015 and have shown themselves to provide a crucial look at the current state of the art in liveness detection schemes. There has been a noticeable increase in the number of participants in LivDet competitions as well as a noticeable decrease in error rates across competitions. Participants have grown from four to the most recent thirteen submissions for Fingerprint Part 1. Fingerprints Part 2 has held steady at two submissions each competition in 2011 and 2013 and only one for the 2015 edition. The continuous increase of competitors demonstrates a growing interest in the topic.
\end{abstract}

\begin{keyword}
Fingerprint \sep Liveness Detection \sep Biometric
%\MSC[2010] 00-01\sep  99-00
\end{keyword}

\end{frontmatter}

\linenumbers

\section{Introduction}
\label{intro}

Among biometrics, fingerprints are probably the best-known and widespread because of the fingerprint properties: universality, durability and individuality. Unfortunately it has been shown that fingerprint scanners are vulnerable to presentation attacks\footnote{Traditionally, a majority of papers refer to these types of attacks as ``spoofing attacks" but recently the term ``presentation attacks" has become the standard \cite{standard}. Similarly, presentation attack detection is the standard term for liveness detection. Presentation attack detection is a more general term and can refer to multiple approaches for detecting a presentation attacks beyond liveness detection.} with an artificial replica of a fingerprint. Therefore, it is important to develop countermeasures to those attacks.

Numerous methods have been proposed to solve the susceptibility of fingerprint devices to attacks by spoof fingers. One primary countermeasure to spoofing attacks is called ``liveness detection'' or presentation attack detection.
Liveness detection is based on the principle that additional information can be garnered above and beyond the data procured and/or processed by a standard verification system, and this additional data can be used to verify if an image is authentic. Liveness detection uses either a hardware-based or software-based system coupled with the authentication program to provide additional security.  Hardware-based systems use additional sensors to gain measurements outside of the fingerprint image itself to detect liveness. Software-based systems use image processing algorithms to gather information directly from the collected fingerprint to detect liveness. These systems classify images as either live or fake.

Since 2009, in order to assess the main achievements of the state of the art in fingerprint liveness detection, University of Cagliari and Clarkson University organized the first Fingerprint Liveness Detection Competition. 

The First International Fingerprint Liveness Detection Competition (LivDet) 2009 \cite{ld09}, provided an initial assessment of software systems based on the fingerprint image only. The second, third and fourth Liveness Detection Competitions (LivDet 2011 \cite{ld11}, 2013 \cite{ld13} and 2015 \cite{ld15}) were created in order to ascertain the progressing state of the art in liveness detection, and also included integrated system testing.

This paper reviews the previous LivDet competitions and how they have evolved over the years. Section 2 of this paper describes the background of spoofing and liveness detection. Section 3 details the methods used in testing for the LivDet competitions as well as descriptions of the datasets that have generated from the competition so far. Section 4 discusses the trends across the competitions reflecting advances in the state of the art. Section 5 concludes the paper and discusses the future of the LivDet competitions.

%In this paper, we describe the four LivDet competitions characteristics and we summarize the results achieved from
%the participants. Section 2 gives a brief overview of fingerprint spoofing techniques and liveness detection countermeasures, describes Data sets and protocols and deepens some specific challenges presented in the two last editions. Experimental results are presented in section 3 (algorithms) and in section 4 (systems).
%Section 5 concludes the paper.

\section{Background}
\label{sec:background}

The concept of spoofing has existed for some time now. Research into spoofing can be seen beginning in 1998 from research conducted by D. Willis and M. Lee where six different biometric fingerprint devices were tested against fake fingers and it was found that four of the six were susceptible to spoofing attacks \cite{sixbd}. This research was approached again in 2000-2002 by multiple institutions including; Putte and Kuening as well as Matsumoto et al. \cite{bfr,iagf}. Putte et al. examined different types of scanning devices as well as different ways of counterfeiting fingerprints \cite{bfr}. The research presented by these researchers looked at the vulnerability of spoofing. In 2001, Kallo (et al.) looked at a hardware solution to Liveness Detection; while in 2002, Schuckers delved into using software approaches for Liveness Detection \cite{fra,sasm}. Liveness detection, with either hardware-based or software-based systems, is used to check if a presented fingerprint originates from a live person or an artificial finger. Usually the result of this analysis is a score used to classify images as either live or fake.

Many solutions have been proposed to solve the vulnerability of spoofing \cite{spfs,fsr}. Bozhao Tan et al. has proposed a solution based on ridge signal and valley noise analysis \cite{spfs}. This solution examines the perspiration patterns along the ridge and the patterns of noise in the valleys of images \cite{spfs}. It was proposed that since live fingers sweat, but spoof fingers do not, the live fingerprint will look ``patchy'' compared to a spoof \cite{spfs}. Also it was proposed that due to the properties of a spoof material, spoof fingers will have granules in the valleys that live fingers will not have \cite{spfs}. Pietro Coli et al. examined static and dynamic features of collected images on a large data set of images \cite{fsr}.

\begin{figure}[t]
\begin{center}
  \includegraphics[width=0.5\textwidth]{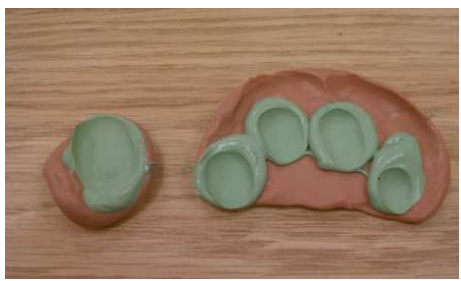}
\end{center}
\caption{Negative impression of five fingers using consensual method.}
\label{fig:mold}       % Give a unique label
\end{figure}

\begin{figure}[t]
\begin{center}
  \includegraphics[width=0.4\linewidth]{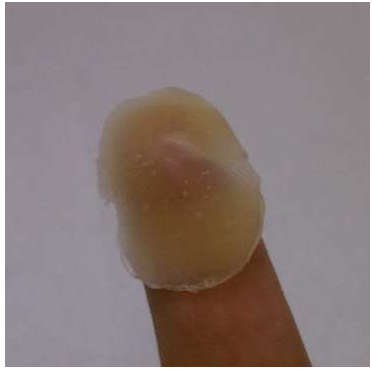}
\end{center}
\caption{Latex spoof on finger.}
\label{fig:fakeFinger}       % Give a unique label
\end{figure}

There are two general forms of creating artificial fingers, the cooperative method and non-cooperative method.  In the cooperative method the subject pushes their finger into a malleable material such as dental impression material, plastic, or wax creating a negative impression of the fingerprint as a mold, see Figure \ref{fig:mold}.  The mold is then filled with a material, such as gelatin, PlayDoh or silicone. This cast can be used to represent a finger from a live subject, see Figure  \ref{fig:fakeFinger}.

\begin{figure}[t]
\begin{center}
  \includegraphics[width=0.6\linewidth]{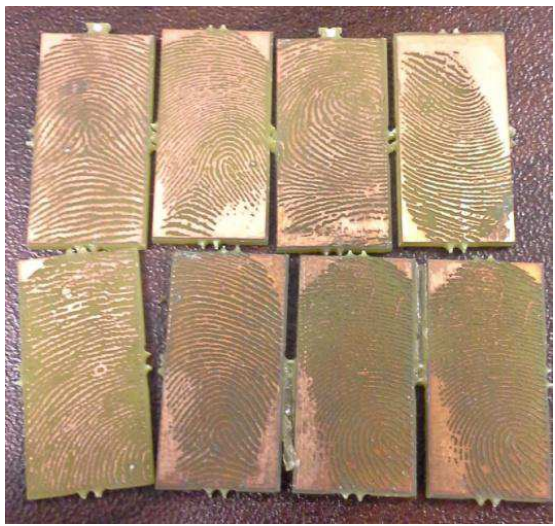}
\end{center}
\caption{Etched fingerprints on PCB.}
\label{fig:latentPCB}       % Give a unique label
\end{figure}

The non-cooperative method involves enhancing a latent fingerprint left on a surface, digitizing it through the use of a photograph, and finally printing the negative image on a transparency sheet. This printed image can then be made into a mold, for example, by etching the image onto a printed circuit board (PCB) which can be used to create the spoof cast as seen on Figure \ref{fig:latentPCB}.

Most competitions focus on matching, such as the Fingerprint Verification Competition held in 2000, 2002, 2004 and 2006 \cite{fvc06} and the ICB Competition on Iris Recognition (ICIR2013) \cite{icbir}. However, these competitions did not consider spoofing.

The Liveness Detection Competition series was started in 2009 and created a benchmark for measuring liveness detection algorithms, similar to matching performance. At that time, there had been no other public competitions held that has examined the concept of liveness detection as part of a biometric modality in deterring spoof attacks.
In order to understand the motivation of organizing such a competition, we observed that the first trials to face with this topic were often carried out with home-made data sets that were not publicly available, experimental protocols were not unique, and the same reported results were obtained on very small data sets. We pointed out these issues in \cite{vitdet}.

Therefore, the basic goal of LivDet has been since its birth to allow researchers testing their own algorithms and systems on publicly available data sets, obtained and collected with the most updated techiques to replicate fingerprints enabled by the experience of Clarkson and Cagliari laboratories, both active on this problem since 2000 and 2003, respectively. At the same time, using a ``competition" instead of simply releasing data sets, could be assurance of a free-of-charge, third-party testing using a sequestered test set. (Clarkson and Cagliari has never took part in LivDet as competitors, due to conflict of interest.)

LivDet 2009 provided results which demonstrated the  state of the art at that time \cite{ld09} for fingerprint systems. LivDet continued in 2011, 2013 and 2015 \cite{ld11,ld13,ld15} and contained two parts: evaluation of software-based systems in Part 1: Algorithms, and evaluation of integrated systems in Part 2: Systems.
Fingerprint will be the focus of this paper. However, LivDet 2013 also included a Part 1: Algorithms for the Iris biometric \cite{ldiris2013} and is continuing in 2015.

Since 2009, evaluation of spoof detection for facial systems was performed in the Competition on Counter Measures to 2-D Facial Spoofing Attacks, first held in 2011 and then held a second time in 2013. The purpose of this competition is to address different methods of detection for 2-D facial spoofing \cite{fsa}. The competition dataset consisted of 400 video sequences, 200 of them real attempts and 200 attack attempts \cite{fsa}. A subset was released for training and then another subset of the dataset was used for testing purposes.

\begin{table*}[!h]
\begin{center}
\begin{tabular}[t]{ | l | r | r | r | r | r | r | r |}
\hline
& \textbf{2010} & \textbf{2011} & \textbf{2012} & \textbf{2013} & \textbf{2014} & \textbf{2015} & \textbf{Sum} \\ \hline\hline
\textbf{LivDet 2009}	 & 4	 & 6	 & 9	 & 2	 & 23	 & 21	 & 65\\ \hline
\textbf{LivDet 2011}	 &  &  & 3	 & 7	 & 21	 & 10	 & 41\\ \hline
\textbf{LivDet 2013}	 &  &  &  & & 16	 & 10	 & 26\\ \hline
\end{tabular}
\caption{Number of LivDet citations on Google Scholar over the years.}
\label{tab:GScite}
\end{center}
\end{table*}

%\section{Short literature review on impact of the LivDet datasets.}
%\label{sec:citations}

During these years many works cited the publications related to the first three LivDet competitions, 2009 \cite{ld09}, 2011 \cite{ld11} and 2013 \cite{ld13}. A quick Google Scholar research produced 65 results for 2009, 41 for 2011 and 26 for 2013. Their distribution is shown in more detail in Table \ref{tab:GScite}, ordered by publication year\footnote{Research was last updated on October 1, 2015}.
In Tables \ref{tab:citation2009}, \ref{tab:citation2011} and \ref{tab:citation2013} a partial list of these publications is presented.

\begin{table*}[!htbp]
\begin{center}
\begin{tabular}[t]{ | p{0.3\linewidth} | p{0.4\linewidth} | p{0.2\linewidth} |}
\hline
\textbf{Authors} & \textbf{Algorithm Type} & \textbf{Performance (Average Classification Error)} \\ \hline\hline
J. Galbally, et al.	\cite{hpfld} &	Quality Related Features.	&	6.6\%\\ \hline
%B. Biggio, et al. \cite{secev} &	N.A.	&	N.A.\\ \hline
E. Marasco, and C. Sansone	\cite{perspmorph}	&	Perspiration and Morphology-based Static Features	&	 12.5\% \\ \hline
J. Galbally, et al.	\cite{iqa}	&	Image Quality Assessment	&	8.2\\ \hline
E. Marasco, and C. Sansone	\cite{mtf}	&	Multiple Textural Features	&	12.5\%\\ \hline
L. Ghiani, et al. \cite{expres}	&	Comparison of Algorithms	&	N.A.\\ \hline
D. Gragnaniello, et al. \cite{wml}	&	Wavelet-Markov Local	&	2.8\%\\ \hline
%P. B. Patil, and H. Shabahat \cite{antist}	&	N.A.	&	N.A.\\ \hline
R. Nogueira, et al. \cite{convnet}	&	Convolutional Networks	&	3.9\% \\ \hline
Y. Jiang, and L. Xin \cite{coomat}	&	Co-occurrence Matrix	&	6.8\%\\ \hline
\end{tabular}
\caption{Publications that cite the LivDet 2009 paper.}
\label{tab:citation2009}
\end{center}
\end{table*}

\begin{table*}[!htbp]
\begin{center}
\begin{tabular}[t]{ | p{0.3\linewidth} | p{0.4\linewidth} | p{0.2\linewidth} |}
\hline
\textbf{Authors} & \textbf{Algorithm Type} & \textbf{Performance (Average Classification Error)} \\ \hline\hline
L. Ghiani, et al. \cite{expres}	&	Comparison of Algorithms	&	N.A.\\ \hline
X. Jia, et al. \cite{mslbp}	&	Multi-Scale Local Binary Pattern	&	7.5\% and 8.9\%\\ \hline
D. Gragnaniello, et al. \cite{lcp}	&	Local Contrast Phase Descriptor	&	5.7\%\\ \hline
N. Poh, et al. \cite{lrc}	&	Likelihood Ratio Computation	&	N.A.\\ \hline
A. F. Sequeira, and J. S. Cardoso \cite{fldpci}	&	Modeling the Live Samples Distribution &	N.A.\\ \hline
%Fumera, Giorgio, et al. \cite{multias}	&	N.A.	&	N.A.\\ \hline
%N. Poh, et al. \cite{symmatch}	&	N.A.	&	N.A.\\ \hline
L. Ghiani, et al.	&	Binarized Statistical Image Features.	&	7.2\%\\ \hline
%Rattani, Ajita, Norman Poh, and Arun Ross	&	A bayesian approach for modeling sensor influence on quality, liveness and match score values in fingerprint verification.	&	IEEE International Workshop on Information Forensics and Security (WIFS 2013)\\ \hline
X. Jia, et al. \cite{msltp}	&	Multi-Scale Local Ternary Patterns	&	9.8\%\\ \hline
%Rattani, Ajita, and Arun Ross.	&	Minimizing the impact of spoof fabrication material on fingerprint liveness detector.	&	IEEE International Conference on Image Processing (ICIP 2014)\\ \hline
G.L. Marcialis, et al. \cite{msltp}	&	Comparison of Algorithms	&	N.A.\\ \hline
R. Nogueira, et al. \cite{convnet}	&	Convolutional Networks	&	6.5\% \\ \hline
Y. Zhang, et al. \cite{walbp}	&	Wavelet Analysis and Local Binary Pattern	&	12.5\%\\ \hline
A. Rattani, et al. \cite{osfsd} &	Textural Algorithms	&	N.A.\\ \hline
P. Johnson, and S. Schuckers \cite{porechar}	&	Pore Characteristics	&	12.0\%\\ \hline
X. Jia, et al. \cite{ocsvm}	&	One-Class SVM	&	N.A.\\ \hline
Y. Jiang, and L. Xin \cite{coomat}	&	Co-occurrence Matrix	&	11.0\%\\ \hline
\end{tabular}
\caption{Publications that cite the LivDet 2011 paper.}
\label{tab:citation2011}
\end{center}
\end{table*}

\begin{table*}[!htbp]
\begin{center}
\begin{tabular}[t]{ | p{0.3\linewidth} | p{0.4\linewidth} | p{0.2\linewidth} |}
\hline
\textbf{Authors} & \textbf{Algorithm Type} & \textbf{Performance (Average Classification Error)} \\ \hline\hline
C. Gottschlich, et al. \cite{hig}	&	Histograms of Invariant Gradients.	&	6.7\%\\ \hline	
R. Nogueira, et al. \cite{convnet}	&	Convolutional Networks	&	3.6\% \\ \hline
Y. Zhang, et al. \cite{walbp}	&	Wavelet Analysis and Local Binary Pattern	&	2.1\%\\ \hline
P. Johnson, and S. Schuckers \cite{porechar}	&	Pore Characteristics	&	N.A.\%\\ \hline
%Wild, Peter, et al.	&	Towards anomaly detection for increased security in multibiometric systems: Spoofing-resistant 1-median fusion eliminating outliers.	&	IEEE International Joint Conference on Biometrics (IJCB 2014)\\ \hline
\end{tabular}
\caption{Publications that cite the LivDet 2013 paper.}
\label{tab:citation2013}
\end{center}
\end{table*}

\section{Methods and Datasets}
\label{sec:methods}

The LivDet competitions feature two distinct parts; Part 1: Algorithms and Part 2: Systems with protocols designed to eliminate variability that may be present across different algorithms or systems. The protocols for each part will be described in further detail in this section with descriptions of each dataset created through this competition.

\subsection{Part 1: Algorithm Datasets}
\label{sec:alg}

The datasets for Part 1: Algorithms changes with each competition. Each competition consists of three to four datasets of live and spoof images from different devices. Eighteen total datasets have been completed and made available thus far in the past four competitions. Fifteen fingerprint liveness datasets and three iris liveness datasets. 

\begin{figure}[t]
\begin{center}
  \includegraphics[width=0.6\linewidth]{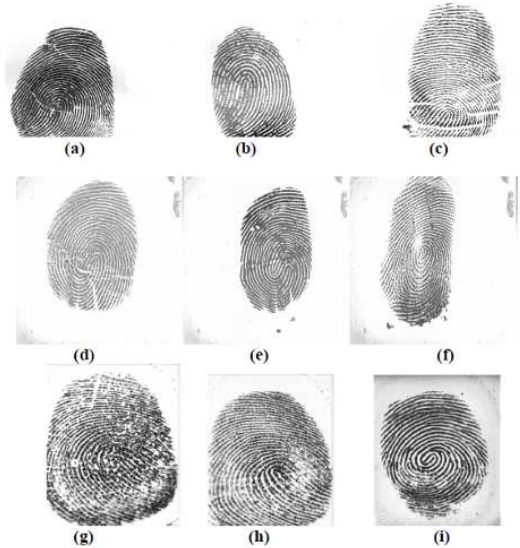}
\end{center}
\caption{Examples of spoof images of the LivDet 2009 datasets. Crossmatch (top): (a) Play-Doh, (b) gelatin, (c) silicone; Identix (middle): (d) Play-Doh, (e) gelatin, (f) silicone; Biometrika (top): (g) Play-Doh, (h) gelatin, (i) silicone.}
\label{fig:spoof09}       % Give a unique label
\end{figure}

LivDet 2009 %featured the starting liveness datasets and 
consisted of data from three optical sensors; Crossmatch, Identix, and Biometrika. The fingerprint images were collected using the consensual approach from three different spoof material types; gelatin, silicone, and play-doh. 
and numbers of images available %are seen in Tables \ref{tab:sensors09}, \ref{tab:train09} and \ref{tab:test09} and more information 
can be found in \cite{ld09}.
Figure \ref{fig:spoof09} shows example images from the datasets.

%\begin{table*}[p]
%\begin{center}
%\begin{tabular}[t]{ | l | l | c | c | c |}
%\hline
%\textbf{Scanner} & \textbf{Model No.} & \textbf{Resolution} & \textbf{Image Size} & %\textbf{Format} \\
% &  & [dpi] & [px] & \\ \hline\hline
%Crossmatch & Verifier 300 LC & 500 & 480x640 & BMP\\ \hline
%Identix & DFR2100 & 686 & 720x720 & TIF\\ \hline
%Biometrika & FX2000 & 569 & 312x372 & TIF\\ \hline
%\end{tabular}
%\caption{Device characteristics for LivDet 2009 datasets.}
%\label{tab:sensors09}
%\end{center}
%\end{table*}

%\begin{table*}[p]
%\begin{center}
%\begin{tabular}[t]{ | l || c | c | c | c |}
%\hline
%\textbf{Dataset} & \textbf{Live} & \textbf{Gelatine} & \textbf{Playdoh} & %\textbf{Silicone}\\ \hline\hline
%Crossmatch & 1000 & 344 & 346 & 310\\ \hline
%Identix & 750 & 250 & 250 & 250\\ \hline\hline
% & \textbf{Live} & \textbf{Silicone} & \textbf{ - } & \textbf{ - } \\ \hline\hline
%Biometrika & 520 & 520 & - & -\\ \hline
%\end{tabular}
%\caption{Number of images for each LivDet 2009 training sets.}
%\label{tab:train09}
%\end{center}
%\end{table*}

%\begin{table*}[p]
%\begin{center}
%\begin{tabular}[t]{ | l || c | c | c | c |}
%\hline
%\textbf{Dataset} & \textbf{Live} & \textbf{Gelatine} & \textbf{Playdoh} & %\textbf{Silicone}\\ \hline\hline
%Crossmatch & 3000 & 1036 & 1034 & 930\\ \hline
%Identix & 2250 & 750 & 750 & 750\\ \hline\hline
% & \textbf{Live} & \textbf{Silicone} & \textbf{ - } & \textbf{ - } \\ \hline\hline
%Biometrika & 1473 & 1480 & - & -\\ \hline
%\end{tabular}
%\caption{Number of images for each LivDet 2009 testing sets.}
%\label{tab:test09}
%\end{center}
%\end{table*}

\begin{figure}[t]
\begin{center}
  \includegraphics[width=0.6\linewidth]{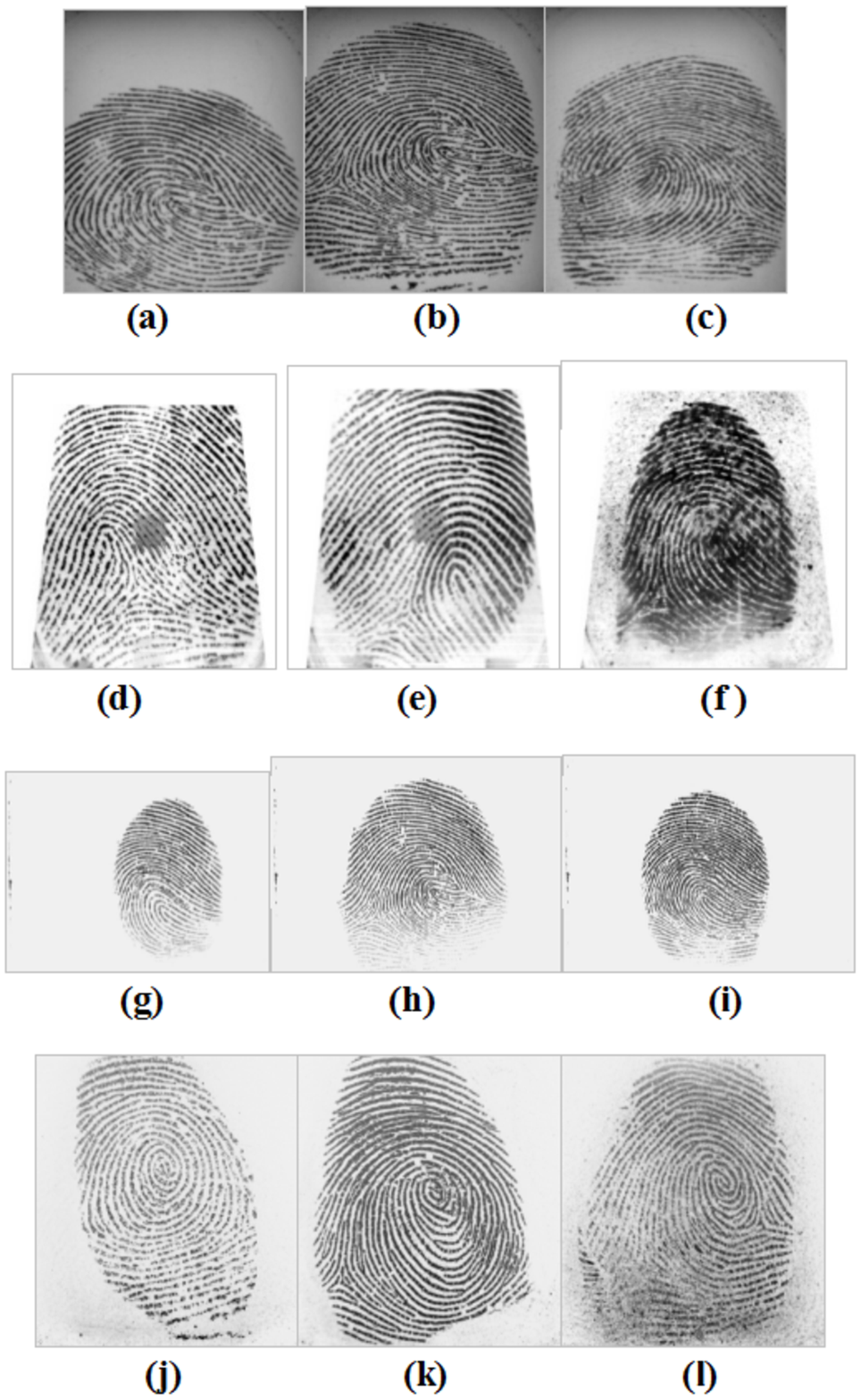}
\end{center}
\caption{Examples of fake fingerprint images of the LivDet 2011 datasets, from Biometrika (a) latex, (b) gelatin, (c) silicone; from Digital Persona: (d) latex, (e) gelatin, (f) silicone; from Italdata: (g) latex (h) gelatin (i) silicone; from Sagem: (j) latex (k) gelatin (l) silicone.}
\label{fig:spoof11}       % Give a unique label
\end{figure}

The dataset for LivDet 2011 consisted of images from four different optical devices, %. Each of these devices has 4000 images split between 2000 live and 2000 spoof images. Both training and testing database consisted of 1000 live and spoof images for each dataset. The devices used for the datasets were 
Biometrika, Digital Persona, ItalData and Sagem. The spoof materials %used for Digital Persona and Sagem were gelatin, latex, Play-doh, silicone, and wood glue. The materials for Biometrika and Italdata were gelatin, latex, ecoflex, silicone, and wood glue. 
were gelatin, latex, ecoflex, Play-doh, silicone and wood glue.
%Tables \ref{tab:sensors11}, \ref{tab:test11} showcase information and numbers of images in the dataset. 
More information can be found in \cite{ld11}. Figure \ref{fig:spoof11} shows images used in the database.

%\begin{table*}[p]
%\begin{center}
%\begin{tabular}[t]{ | l | l | c | c | c |}
%\hline
%\textbf{Scanner} & \textbf{Model} & \textbf{Resolution} & \textbf{Image Size} & %\textbf{Format} \\ 
% &  & [dpi] & [px] &  \\ \hline\hline
%Biometrika & FX2000 & 569 & 312x372 & PNG\\ \hline
%Digital Persona & 4000B & 500 & 355x391 & BMP\\ \hline
%Italdata & ET10 & 500 & 640x480 & PNG\\ \hline
%Sagem & MSO300 & 500 & 352x384 & BMP\\ \hline
%\end{tabular}
%\caption{Device characteristics for LivDet 2011 datasets.}
%\label{tab:sensors11}
%\end{center}
%\end{table*}

%\begin{table*}[p]
%\begin{center}
%\begin{tabular}{ | l || c | c | c | c | c | c |}
%\hline
%\textbf{Dataset} & \textbf{Live} & \textbf{Ecoflex} & \textbf{Gela} & \textbf{Latex} & %\textbf{Silgum} & \textbf{Wood} \\
% &  &  & \textbf{tine} &  &  & \textbf{Glue} \\ \hline\hline
%Biometrika & 1000 & 200 & 200 & 200 & 200 & 200\\ \hline
%Italdata & 1000 & 200 & 200 & 200 & 200 & 200\\ \hline
% & \textbf{Live} & \textbf{Playdoh} & \textbf{Gela} & \textbf{Latex} & \textbf{Sili} & %\textbf{Wood} \\
% &  &  & \textbf{tine} &  & \textbf{cone} & \textbf{Glue} \\ \hline\hline
%Digital P. & 1000 & 200 & 200 & 200 & 200 & 200\\ \hline\hline
%Sagem & 1000 & 200 & 200 & 200 & 200 & 200\\ \hline
%\end{tabular}
%\caption{Number of images for each LivDet 2011 testing sets. Training sets were a similar %size.}
%\label{tab:test11}
%\end{center}
%\end{table*}

\begin{figure}[t]
\begin{center}
  \includegraphics[width=1\linewidth]{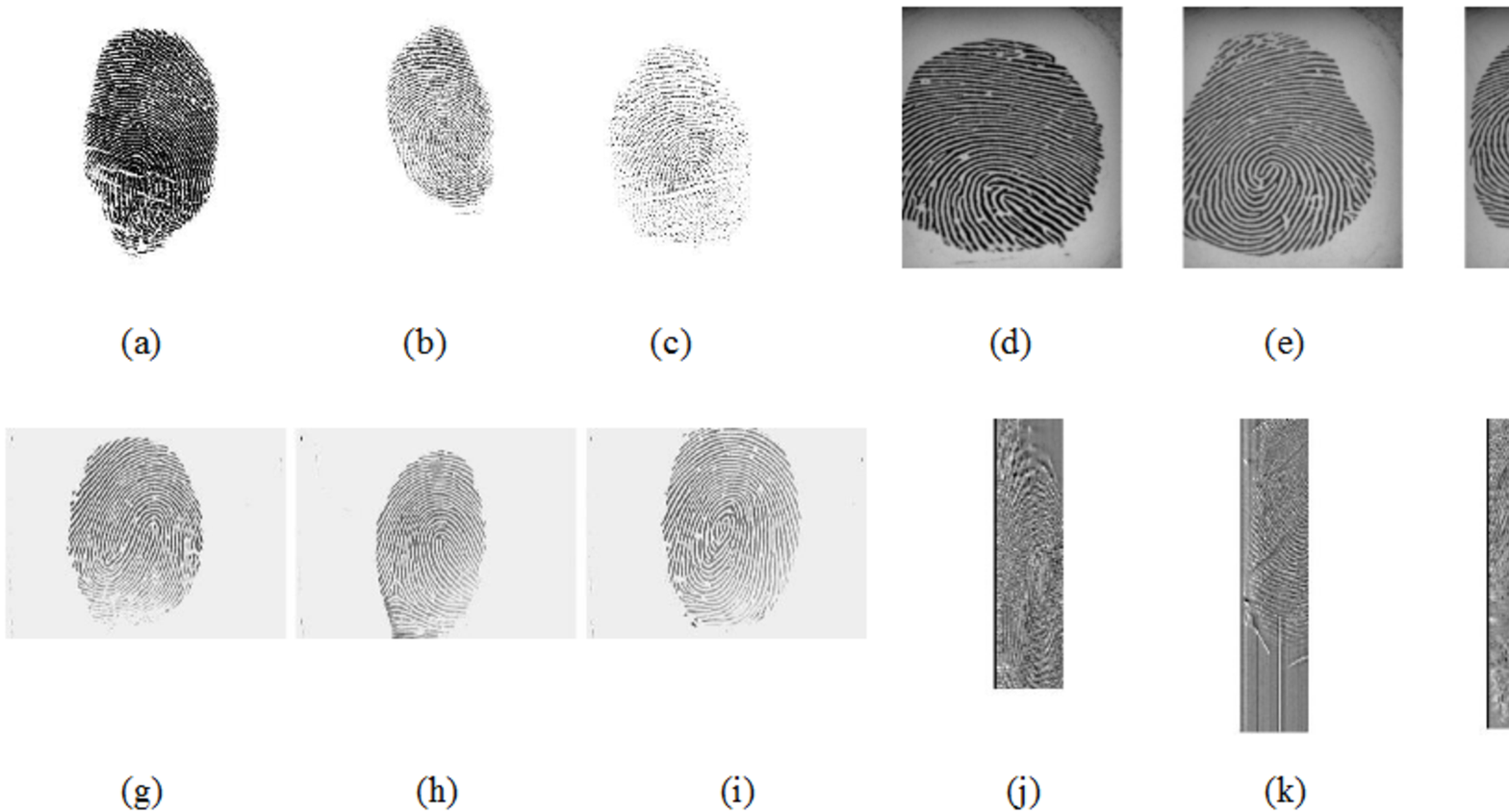}
\end{center}
\caption{Examples of fake fingerprint images of the LivDet 2013. From Crossmatch (a) body double, (b) latex, (c) wood glue, from Biometrika (d) gelatine, (e) latex, (f) wood glue, from Italdata (g) gelatine, (h) latex, (i) wood glue, from Swipe (j) body double, (k) latex, (l) wood glue.}
\label{fig:spoof13}       % Give a unique label
\end{figure}

The dataset for LivDet 2013 consisted of images from four different devices; Biometrika, Crossmatch, ItalData and Swipe. %Biometrika and ItalData had a total of 4000 images, 2000 of which were spoof made from gelatin, latex, ecoflex, modasil, and wood glue. Crossmatch and Swipe had 4500 images, 2000 of which were made from body double, latex, play-doh, and wood glue. 
Spoofs were made from gelatin, body double, latex, play-doh, ecoflex, modasil, and wood glue.
LivDet 2013 featured the first use of the non-cooperative method for creating spoof images and was used for Biometrika and ItalData. % Tables \ref{tab:sensors13} and \ref{tab:test13} covers the device and dataset characteristics for LivDet 2013 Fingerprint Part 1 and 
More information can be found in \cite{ld13}. Figure \ref{fig:spoof13} gives example images from the databases.

%\begin{table*}[p]
%\begin{center}
%\begin{tabular}[t]{ | l | l | c | c | c |}
%\hline
%\textbf{Scanner} & \textbf{Model} & \textbf{Resolution} & \textbf{Image Size} & \textbf{Format} \\
% &  & [dpi] & [px] &  \\ \hline\hline
%Biometrika & FX2000 & 569 & 312x372 & PNG\\ \hline
%Italdata & ET10 & 500 & 640x480 & PNG\\ \hline
%Crossmatch & L Scan Guardian & 500 & 800x750 & BMP\\ \hline
%Swipe &  & 96 & 208x1500 & BMP\\ \hline
%\end{tabular}
%\caption{Device characteristics for LivDet 2013 datasets.}
%\label{tab:sensors13}
%\end{center}
%\end{table*}

%\begin{table*}[p]
%\begin{center}
%\begin{tabular}[t]{ | l || c | c | c | c | c | c |}
%\hline
%\textbf{Dataset} & \textbf{Live} & \textbf{Ecoflex} & \textbf{Gela} & \textbf{Latex} & \textbf{Silgum} & \textbf{Wood} \\
% &  &  & \textbf{tine} &  &  & \textbf{Glue} \\ \hline\hline
%Biometrika & 1000 & 200 & 200 & 200 & 200 & 200\\ \hline
%Italdata & 1000 & 200 & 200 & 200 & 200 & 200\\ \hline\hline
% & \textbf{Live} & \textbf{Body} & \textbf{Latex} & \textbf{Playdoh} & \textbf{Wood} & \textbf{ - } \\
% &  & \textbf{Double} &  &  & \textbf{Glue} & \textbf{ - } \\ \hline\hline
%Crossmatch & 1250 & 250 & 250 & 250 & 250 & -\\ \hline
%Swipe & 1153 & 250 & 250 & 250 & 250 & -\\ \hline
%\end{tabular}
%\caption{Number of images for each LivDet 2013 testing sets. Training sets were a similar %size.}
%\label{tab:test13}
%\end{center}
%\end{table*}

%LivDet 2013 also contained a Part 1: Algorithms for the Iris biometric. This database used three different sensors to create datasets. The spoof images that were collected came from one of two different presentation attack types; patterned contact lenses being worn by a human subject to obscure the natural iris pattern, and the second is printed iris images on paper to identify as another person. The number of images used in each dataset is shown in Table \ref{fig:irisDB} with some live and spoof images shown in figure \ref{fig:irisLiveSpoof}.
%
%\begin{figure}[t]
%\begin{center}
%  \includegraphics[width=0.5\linewidth]{irisDatabase.eps}
%\end{center}
%\caption{Image Iris database.}
%\label{fig:irisDB}       % Give a unique label
%\end{figure}
%
%\begin{figure}[t]
%\begin{center}
%  \includegraphics[width=1\linewidth]{irisLiveSpoof.eps}
%\end{center}
%\caption{Examples of live and spoof images.}
%\label{fig:irisLiveSpoof}       % Give a unique label
%\end{figure}

The dataset for LivDet 2015 consists of images from four different optical devices; Green Bit, Biometrika, Digital Persona and Crossmatch. %The detailed characteristics of the sensors are shown in Table \ref{tab:sensors15}.
%For each of these devices there are more than 4000 images. Live images came from multiple acquisitions of all fingers of different subjects. Each finger was acquired in a variety of ways in order to mimic real scenarios.  Acquisitions include normal mode, with wet and dry fingers and with high and low pressure. The spoof images of the LivDet 2015 datasets were collected using the cooperative method that was earlier described.
The spoof materials were Ecoflex, gelatin, latex, wood glue, a liquid Ecoflex and RTV (a two-component silicone rubber) for the Green Bit, the Biometrika and the Digital Persona datasets, and Playdoh, Body Double, Ecoflex, OOMOO (a silicone rubber) and a novel form of gelatin for Crossmatch dataset. %The entire datasets were divided into two parts by using images from different subjects: a training set, for the configuration of algorithms, and a testing set to evaluate the performance. The testing sets included spoof images of unknown materials, i.e. materials which were not included in the training set. The unknown materials are liquid Ecoflex and RTV for Green Bit, Biometrika and Digital Persona datasets, and OOMOO and Gelatin for Crossmatch dataset.
%Tables \ref{tab:sensors15} and \ref{tab:test15} covers the device and dataset characteristics, m
More information can be found in \cite{ld15}.

\subsection{Part 2: Systems Submissions}
\label{sec:syst}

Public datasets were not released from systems collections, however data was collected on the submitted systems. Unlike in Part 1: Algorithms where data was pre-generated before the competition, Part 2: Systems data was collected through systematic testing of submitted system. For LivDet 2011 this consisted of 500 live attempts from 50 people (totaling 5 images for each of the R1 and R2 fingers) as well as 750 attempts with spoofs of five materials (play-doh, gelatin, silicone, body double, and latex).  For LivDet 2013, 1000 live attempts were conducted as well as 1000 spoof attempts from the materials; Play-Doh, gelatin, Ecoflex, Modasil, and latex.
In 2015 the system was tested using the three known spoof recipes. Two unknown spoof recipes were also tested to examine the flexibility of the sensor toward novel spoof methods. The known recipes were Playdoh, Body Double, and Ecoflex. The two unknown recipes used were OOMOO (a silicone rubber) and a novel form of gelatin. 2011 attempts were completed with 1010 live attempts from 51 subjects (2 images each of all 10 fingers) and 1001 spoof attempts across the five different materials giving approximately 200 images per spoof type. 500 spoofs were created from each of 5 fingers of 20 subjects for each of the five spoof materials. Two attempts were performed with each spoof.

The submitted system needs to be able to output a file with the collected image as well as a liveness score on the range of 0 to 100 with 100 being the maximum degree of liveness and 50 being the threshold value to determine if an image is live or spoof. If the system is not able to process a live subject it is counted as a failure to enroll and counted against the performance of the system (as part of Ferrlive). However, if the system is unable to process a spoof finger it is considered as a fake non-response and counted as a positive in terms of system effectiveness for spoof detection.
% (as part of \textit{Ferrspoof}).

\subsection{Image Quality}
\label{sec:quality}

Fingerprint image quality has a powerful effect on the performance of a matcher. Many commercial fingerprint systems contain algorithms to ensure that only higher quality images are accepted to the matcher. This rejects low quality images where low quality images have been shown to degrade the performance of a matcher \cite{nist}. The algorithms and systems submitted for this competition did not use a quality check to determine what images would proceed to the liveness detection protocols. Through taking into account the quality of the images before applying liveness detection a more realistic level of error can be shown.

Our methodology uses the NIST Fingerprint Image Quality (NFIQ) software to examine the quality of all fingerprints used for the competition and examine the effects of removing lower quality fingerprint images on the liveness detection protocols submitted.  NFIQ computes a feature vector from a quality image map and minutiae quality statistics as an input to a multi-layer perceptron neural network classifier \cite{nist}. The quality of the fingerprint is determined from the neural network output. The quality for each image is assigned on a scale from 1 (highest quality) to 5 (lowest quality).

\subsection{Performance Evaluation}
\label{sec:perfEval}

The parameters adopted for the performance evaluation are the following:
%\begin{itemize}
%\item{\textit{Evaluation per sensor/system:}}
%\begin{itemize}
%\item{\textit{Frej\_n}: Rate of failure to enroll for the sub-set \textit{n}.}
%\item{\textit{Frej\_n}: Rate of failure to enroll for the sub-set \textit{n}.}
%\item{\textit{Fcorrlive\_n}: Rate of correctly classified live fingerprints for sub-set \textit{n}.}
%\item{\textit{Fcorrfake\_n}: Rate of correctly classified fake fingerprints for sub-set \textit{n}.}
%\item{\textit{Ferrlive\_n}: Rate of misclassified live fingerprints for sub-set \textit{n}.}
%\item{\textit{Ferrfake\_n}: Rate of misclassified fake fingerprints for sub-set \textit{n}.}
%\item{\textit{FakeNonResponse}: Rate of failure to acquire for fake fingerprints for system.}
%\end{itemize}
%\item{\textit{Overall evaluation:}}
\begin{itemize}
%\item{\textit{Frej}: Rate of failure to enroll.}
%\item{\textit{Fcorrlive}: Rate of correctly classified live fingerprints.}
%\item{\textit{Fcorrfake}: Rate of correctly classified fake fingerprints.}
\item{\textit{Ferrlive}: Rate of misclassified live fingerprints.}
\item{\textit{Ferrfake}: Rate of misclassified fake fingerprints.}
\item{\textit{Average Classification Error (ACE)}: 
$ACE = (\frac{Ferrlive + Ferrfake}{2})$}
\item{\textit{Equal Error Rate (EER)}: Rate at which Ferrlive and Ferrfake are equal.}
\item{\textit{Accuracy}: Rate of correctly classified live and fake fingerprints at a 0.5 threshold.}
\end{itemize}
%\end{itemize}

\subsection{Specific challenges}
\label{sec:partecipants}

In the last two editions of the competition specific challenges were introduced. Two of the 2013 datasets, unlike all the other cases, contain spoofs that were collected using latent fingerprints.  The 2015 edition had two new components:  (1) the testing set included images from two kinds of spoof materials which were not present in the training set in order to test the robustness of the algorithms with regard to unknown attacks, and (2) one of the data sets was collected using a 1000 dpi sensor.

\subsubsection{LivDet 2013 – Consensual vs. Semi-Consensual}
\label{sec:consUncons}

In the consensual method the subject pushed his finger into a malleable material such as silicon gum creating a negative impression of the fingerprint as a mold. The mold was then filled with a material, such as Gelatin.
The ``semi-consensual method" consisted of enhancing a latent fingermark pressed on a surface, and digitizing it through the use of a common scanner.\footnote{Obviously all subjects were fully aware of this process, and gave the full consent to replicate their fingerprints from their latent marks.} Then, through a binarization process and with an appropriate threshold choice, the binarized image of the fingerprint was obtained. The thinning stage allowed the line thickness to be reduced to one pixel obtaining the skeleton of the fingerprint negative. This image was printed on a transparency sheet, in order to have the mold. A gelatin or silicone material was dripped over this image, and, after solidification, separated and used as a fake fingerprint.

The consensual method leads to an almost perfect copy of a live finger, whose mark on a surface is difficult to recognize as a fake unless through an expert dactiloscopist. On the other hand, the spoof created by semi- or unconsensual method is much less similar. In a latent fingerprint, many details are lost and the skeletonization process further deteriorates the spoof quality making it easier to distinguish a live from a fake.
However, while it could be hard to convince someone to leave the cast of a finger, it's potentially much easier to obtain one of his latent fingerprints.
The spoof images in the Biometrika and Italdata 2013 datasets were created by printing the negative image on a transparency sheet. As we will see in the next section, the error rates, as would be expected, are lower than those of the other datasets.

\subsubsection{LivDet 2015 – Hidden Materials and 500 vs 1000 dpi}
\label{sec:hiddenDpi}

As already stated, the testing sets of LivDet 2015 included spoof images of never-seen-before materials. These materials were liquid Ecoflex and RTV for Green Bit, Biometrika and Digital Persona datasets, and OOMOO and Gelatin for Crossmatch dataset. Our aim was to assess the reliability of algorithms. As a matter of fact, in a realistic scenario, the material used to attack a biometric system could be considered unknown as a liveness detector should be able to deal with any kind of spoof material.

Another peculiarity of the 2015 edition was the presence of the Biometrika HiScan-PRO, a sensor with a resolution of 1000 dpi instead of $\sim$500 dpi resolution for most of the datasets used so far in the competition.
It is reasonable to hypothesize that doubling the image resolution, the feature extraction phase should benefit as well as the final performance. The results that we will show in the next section does not confirm this hypothesis.

\section{Examination of Results}
\label{sec:exam}

In this Section, we analyze the experimental results for the four LivDet editions. Results show the growth and improvement across the four competitions.

\subsection{Trends of Competitors and Results for Fingerprint Part 1: Algorithms}
\label{sec:trends1}

The number of competitors for Fingerprint Part 1: Algorithms have increased during the last years. LivDet 2009 contained a total of 4 algorithm submissions. LivDet 2011 saw a slight decrease in competitors with only 3 organizations submitting algorithms, however LivDet 2013 and 2015 gave rise to the largest of the competitions with 11 submitted algorithms, nine participants in the former and ten in the latest. Submissions for each LivDet are detailed in Table \ref{tab:participants1}.

\begin{table*}[p]
\begin{center}
\begin{tabular}[t]{ | c | c |}
\hline
\textbf{Participants LivDet 2009} & \textbf{Algorithm Name} \\ \hline\hline
Dermalog Identification Systems GmbH & Dermalog\\ \hline
Universidad Autonoma de Madrid & ATVS\\ \hline
Anonymous & Anonymous\\ \hline
Anonymous2 & Anonymous2\\ \hline\hline

\textbf{Participants LivDet 2011} & \textbf{Algorithm Name} \\ \hline\hline
Dermalog Identification Systems GmbH & Dermalog\\ \hline
University of Naples Federico II & Federico\\ \hline
Chinese Academy of Sciences & CASIA\\ \hline\hline

\textbf{Participants LivDet 2013} & \textbf{Algorithm Name} \\ \hline\hline
Dermalog Identification Systems GmbH & Dermalog\\ \hline
Universidad Autonoma de Madrid & ATVS\\ \hline
HangZhou JLW Technology Co Ltd & HZ-JLW\\ \hline
Federal University of Pernambuco & Itautec\\ \hline
Chinese Academy of Sciences & CAoS\\ \hline
University of Naples Federico II (algorithm 1) & UniNap1\\ \hline
University of Naples Federico II (algorithm 2) & UniNap2\\ \hline
University of Naples Federico II (algorithm 3) & UniNap3\\ \hline
First Anonymous participant & Anonym1\\ \hline
Second Anonymous participant & Anonym2\\ \hline
Third Anonymous participant & Anonym3\\ \hline\hline

\textbf{Participants LivDet 2015} & \textbf{Algorithm Name} \\ \hline\hline
Instituto de Biociencias, Letras e Ciencias Exatas & COPILHA\\ \hline
Institute for Infocomm Research (I2R) & CSI \\ \hline
Institute for Infocomm Research (I2R) & CSI\_MM \\ \hline
Dermalog & hbirkholz \\ \hline
Universidade Federal de Pernambuco & hectorn \\ \hline
Anonymous participant & anonym \\ \hline
Hangzhou Jinglianwen Technology Co., Ltd & jinglian \\ \hline
Universidade Federal Rural de Pernambuco & UFPE I \\ \hline
Universidade Federal Rural de Pernambuco & UFPE II \\ \hline
University of Naples Federico II & unina\\ \hline
New York University & nogueira \\ \hline
Zhejiang University of Technology & titanz \\ \hline

\end{tabular}
\caption{Participants for Part 1: Algorithms.}
\label{tab:participants1}
\end{center}
\end{table*}

This increase of participants has shown the grown of interest in the topic, which has been coupled with the general decrease of the error rates.

First of all, the two best algorithms for each competition, in terms of performance, are detailed in Table \ref{tab:bestErr} based on the average error rate across the datasets where ``Minimum Average" error rates are the best results and ``Second Average" are the second best results.

\begin{table*}[p]
\begin{center}
\begin{tabular}[t]{ | p{2.3cm} | p{2.3cm} | p{2.3cm} | p{2.3cm} |}
\hline
\multicolumn{4}{|c|}{\textbf{2009}} \\ \hline
\textbf{Minimum Avg Ferrlive} & \textbf{Minimum Avg Ferrfake} & \textbf{Second Avg Ferrlive} & \textbf{Second Avg Ferrfake} \\ \hline
13.2\% & 5.4\% & 20.1\% & 9.0\% \\ \hline\hline

\multicolumn{4}{|c|}{\textbf{2011}} \\ \hline
\textbf{Minimum Avg Ferrlive} & \textbf{Minimum Avg Ferrfake} & \textbf{Second Avg Ferrlive} & \textbf{Second Avg Ferrfake} \\ \hline
11.8\% & 24.8\% & 24.5\% & 24.8\% \\ \hline\hline

\multicolumn{4}{|c|}{\textbf{2013}} \\ \hline
\textbf{Minimum Avg Ferrlive} & \textbf{Minimum Avg Ferrfake} & \textbf{Second Avg Ferrlive} & \textbf{Second Avg Ferrfake} \\ \hline
11.96\% & 1.07\% & 17.64\% & 1.10\% \\ \hline\hline

\multicolumn{4}{|c|}{\textbf{2015}} \\ \hline
\textbf{Minimum Avg Ferrlive} & \textbf{Minimum Avg Ferrfake} & \textbf{Second Avg Ferrlive} & \textbf{Second Avg Ferrfake} \\ \hline
5.13\% & 2.79\% & 6.45\% & 4.26\% \\ \hline

\end{tabular}
\caption{Two best error rates for each competition. It can be noticed a positive trend in terms of both Ferrlive and Ferrfake parameters. In particular 2011 and 2015 exhibited very difficult tasks due to the high quality of fingerprint images thus they should be taken into account as a reference of current liveness detector performance against the ``worst scenario", that is, the high quality reproduction of a subject's fingerprint.}
\label{tab:bestErr}
\end{center}
\end{table*}

There is a stark difference between the results seen from LivDet 2009 to LivDet 2015. LivDet 2009 to LivDet 2011 did not see much decrease in error, where LivDet 2013 and LivDet 2015 each decreased in error from the previous compeition.

The mean values of the ACE (Average Classification Error) over all the participants calculated for each dataset confirm this trend. Mean and standard deviation are shown in Table \ref{tab:meanstd}.

\begin{table*}[p]
\begin{center}
\begin{tabular}[t]{ | l | c | c |}
\hline
\multicolumn{3}{|c|}{\textbf{2009}} \\ \hline
& \textbf{Mean} & \textbf{Std. Dev.} \\ \hline
Identix	& 8.27    & 4.65\\ \hline
Crossmaatch	& 15.59    & 5.60\\ \hline
Biometrika	& 32.59    & 9.64\\ \hline
%\end{tabular}
%\caption{Mean and standard deviation ACE values for LivDet 2009 datasets.}
%\label{tab:meanstd2009}
%\end{center}
%\end{table*}
%
%\begin{table*}[!t]
%\begin{center}
%\begin{tabular}[t]{ | l | c | c |}
\hline
\multicolumn{3}{|c|}{\textbf{2011}} \\ \hline
& \textbf{Mean} & \textbf{Std. Dev.} \\ \hline\hline
Biometrika & 31.30   & 10.25\\ \hline
ItalData & 29.50    & 9.42\\ \hline
Sagem & 16.70    & 5.33\\ \hline
Digital Persona & 23.47   & 13.70\\ \hline
%\end{tabular}
%\caption{Mean and standard deviation ACE values for LivDet 2011 datasets.}
%\label{tab:meanstd2011}
%\end{center}
%\end{table*}
%
%\begin{table*}[!t]
%\begin{center}
%\begin{tabular}[t]{ | l | c | c |}
\hline
\multicolumn{3}{|c|}{\textbf{2013}} \\ \hline
& \textbf{Mean} & \textbf{Std. Dev.} \\ \hline\hline
Biometrika	& 7.32    & 8.80\\ \hline
Italdata	& 12.25   & 18.08\\ \hline
%Crossmatch	& 46.75    & 6.27\\ \hline
Swipe	& 16.67   & 15.30\\ \hline
%\end{tabular}
%\caption{Mean and standard deviation ACE values for LivDet 2013 datasets.}
%\label{tab:meanstd2013}
%\end{center}
%\end{table*}
%
%\begin{table*}[!t]
%\begin{center}
%\begin{tabular}[t]{ | l | c | c |}
\hline
\multicolumn{3}{|c|}{\textbf{2015}} \\ \hline
& \textbf{Mean} & \textbf{Std. Dev.} \\ \hline\hline
GreenBit	& 11.47    & 7.10\\ \hline
Biometrika	& 15.81    & 7.12\\ \hline
DigitalPersona	& 14.89   & 13.72\\ \hline
Crossmatch	& 14.65   & 10.28\\ \hline
\end{tabular}
\caption{Mean and standard deviation ACE values for each dataset of the competition.}
%\caption{Mean and standard deviation ACE values for LivDet 2015 datasets.}
\label{tab:meanstd}
\end{center}
\end{table*}

The standard deviation values ranges between 5 and 18\% depending on the dataset and competition editions. Mean ACE values confirm the error increase in 2011 due to the high quality of cast and fake materials. The low values in 2013 for the Biometrika and Italdata are due, as stated before, to the use of latent fingerprints in the spoof creation process, creating lower quality spoofs that are easier to detect. In order to confirm that, we compared these values with those obtained for the same sensors in 2011 (see Table \ref{tab:11vs13}). Last two rows of Table \ref{tab:11vs13} report average classification error and related standard deviation over above sets. Obviously, further and independent experiments are needed because participants of 2011 and 2013 were different so that different algorithms are likely also. However,  results highlight the performance virtually achievable over two scenarios: a sort of ``worst case", namely, the one represented by LivDet 2011, where quality of spoofs is very high, and a sort of ``realistic case" (LivDet 2013), where spoofs are created from latent marks as one may expect. The fact that even in this case the average error is $10\%$, whilst the standard deviation does not differ with regard to LivDet 2011, should not be underestimated. The improvement could be likely due to the different ways of creating the fakes.

\begin{table*}[p]
\begin{center}
\begin{tabular}[t]{ | l | c | c |}
\hline
& \textbf{Consensual} & \textbf{Semi-consensual} \\
& \textbf{(LivDet 2011)} & \textbf{(LivDet 2013)} \\ \hline\hline
Biometrika & 31.30   & 7.32\\ \hline
ItalData & 29.50    & 12.25\\ \hline\hline
Mean & 30.40   & 9.78\\ \hline
Standard Deviation & 1.27    & 3.49\\ \hline
\end{tabular}
\caption{Comparison between mean ACE values for Biometrika and Italdata datasets from LivDet 2011 and 2013.}
\label{tab:11vs13}
\end{center}
\end{table*}
\clearpage

\begin{figure}[t]
\begin{center}
 \includegraphics[width=0.6\linewidth]{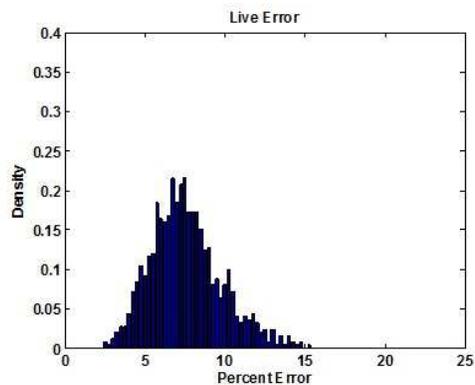}
\end{center}
\caption{FerrLive Rates across 1000 tests for Crossmatch 2013.}
\label{fig:FerrLiveCross} % Give a unique label
\end{figure}

\begin{figure}[t]
\begin{center}
\includegraphics[width=0.6\linewidth]{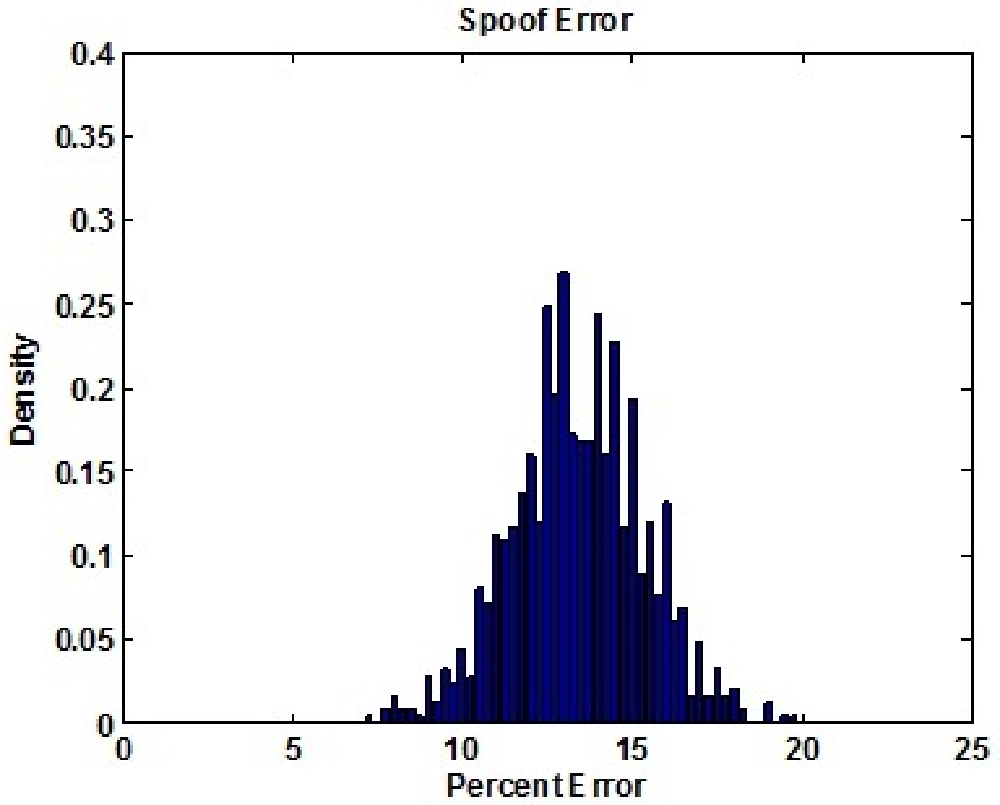}
\end{center}
\caption{FerrFake Rates across 1000 tests for Crossmatch 2013.}
\label{fig:FerrFakeCross} % Give a unique label
\end{figure}

The abnormally high values for the LivDet 2013 Crossmatch dataset occurred due to an occurrence in the Live data. The Live images were difficult for the algorithms to recognize. All data was collected in the same time frame and data in training and testing sets were determined randomly among the data collected. A follow-up test was conducted using benchmark algorithms at University of Cagliari and Clarkson University which revealed similar scores on the benchmark algorithms as the submitted algorithms with initial results had an EER of 41.28\%. 
The data was further tested with 1000 iterations of train/test dataset generation using a random selection of images for the live training and test sets (with no common subjects between the sets which is a requirement for all LivDet competitions). This provided new error rates shown (see Table \ref{tab:1000tests} and Figures  \ref{fig:FerrLiveCross} and \ref{fig:FerrFakeCross})

\begin{table*}[!htbp]
\begin{center}
\begin{tabular}{|c | c | c |} 
\hline
& Average Error Rate & Standard Deviation\\ 
\hline
FerrLive & 7.57\% & 2.21\%\\ 
\hline
FerrFake & 13.41\% & 1.95\%\\ 
\hline
Equal Error Rate & 9.92\% & 1.42\%\\
\hline
\end{tabular}
\caption{Crossmatch 2013 Error Rates across 1000 Tests}
\label{tab:1000tests}
\end{center}
\end{table*}

Examining these results allows us to draw the conclusion that the original selection of subjects was an anomaly that caused improper error rates because each other iteration, even only changing a single subject, dropped FerrLive error rates to 15\% and below. Data is being more closely examined in future LivDet competitions in order to counteract this problem with data being processed on a benchmark algorithm before being given to participants. The solution to this for the LivDet 2013 data going forward is to rearrange the training and test sets for future studies using this data. Researchers will need to be clear which split of training/test they used in their study.
For this reason we removed from the experimental results of those obtained with the Crossmatch 2013 dataset.

\begin{figure}[t]
\begin{center}
  \includegraphics[width=0.9\linewidth]{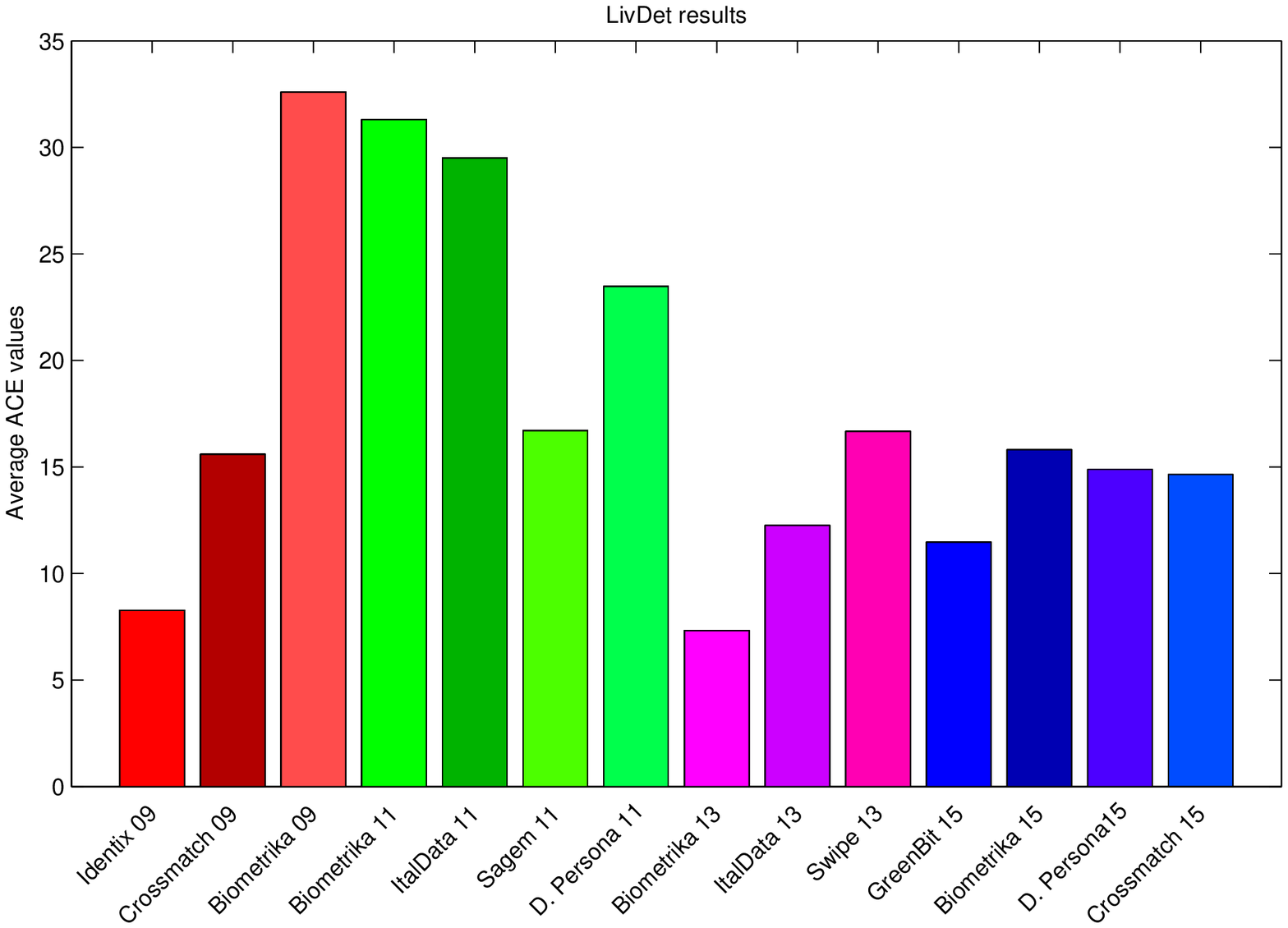}
\end{center}
\caption{LivDet results over the years: near red colors for 2009, near green for 2011, near magenta for 2013 and near blue for 2015.}
\label{fig:barResult}
\end{figure}

LivDet 2015 error rates confirm a decreasing trend with respect to attacks made up of high quality spoofs, with all the mean values between 11\% and 16\%.
These results are summarized in Figure \ref{fig:barResult}.
%Examining the results from Part 2: Systems shows that overall errors have dropped from 2011. The 2013 results were spectacular with a 0\% FerrFake and 1.4\% Ferrlive. While not as high as 2013, the results in 2015 are still an improvement in the field compared to the past. These results are summarized in Figure \ref{fig:sysError}.

%\begin{figure}[t]
%\begin{center}
%  \includegraphics[width=0.8\linewidth]{errorbyyear.eps}
%\end{center}
%\caption{LivDet system results over the years.}
%\label{fig:sysError}
%\end{figure}

Reported error rates suggest a slow, but steady advancement in the art of liveness and artifact detection. This gives supporting evidence that the technology is evolving and learning to adapt and overcome the presented challenges.

Comparing the performance of the LivDet 2015 datasets (as shown in Table \ref{tab:meanstd} and in more details in \cite{ld15}) two other important remarks can be made: the higher resolution for Biometrika sensor did not necessarily achieve the best classification performance, while the small size of the images for the Digital Persona device generally degrades the accuracy of all algorithms.

\begin{figure}[t]
\centering
\subfloat[]{\includegraphics[width=0.45\linewidth]{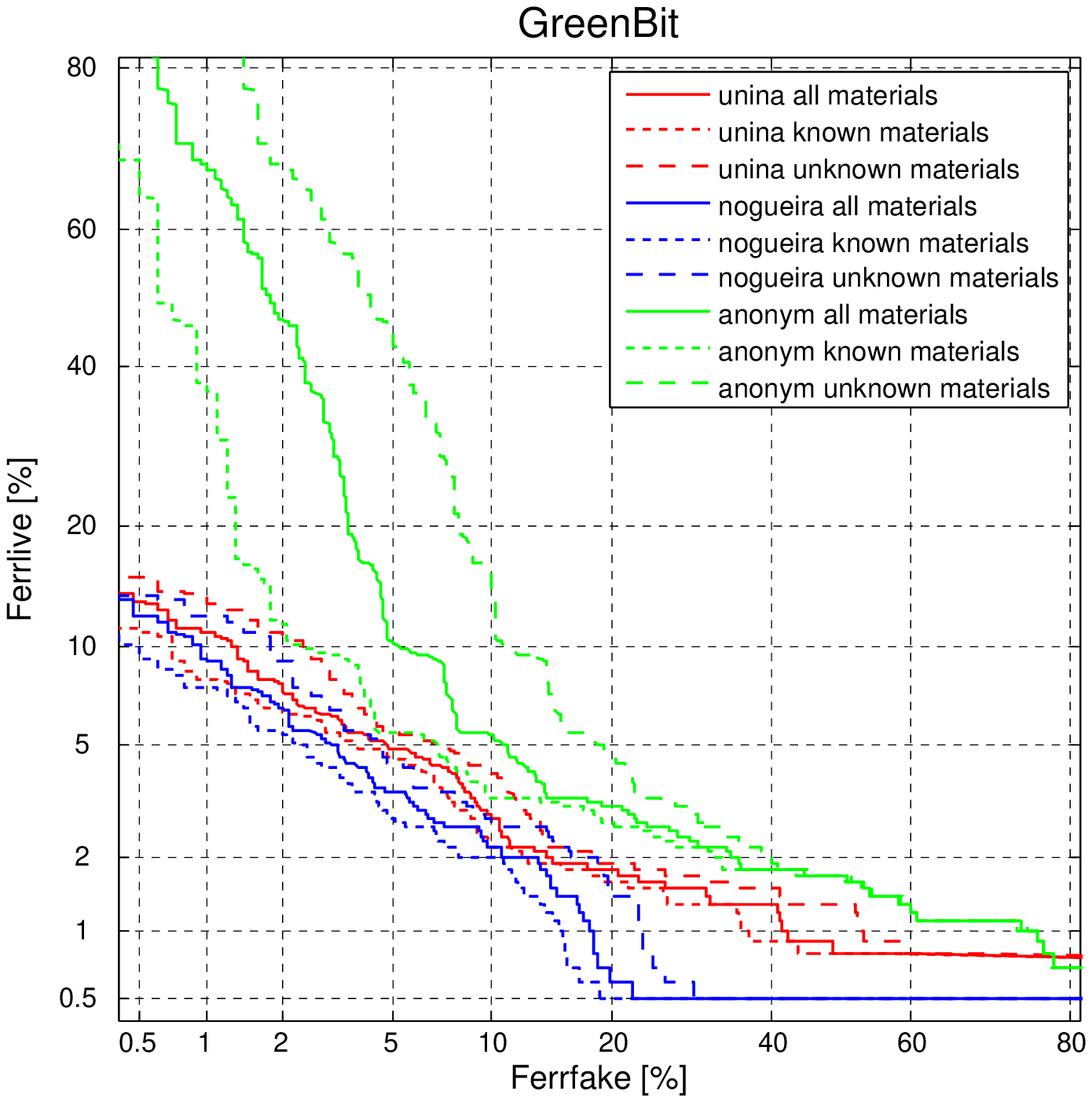}}
\qquad
\subfloat[]{\includegraphics[width=0.45\linewidth]{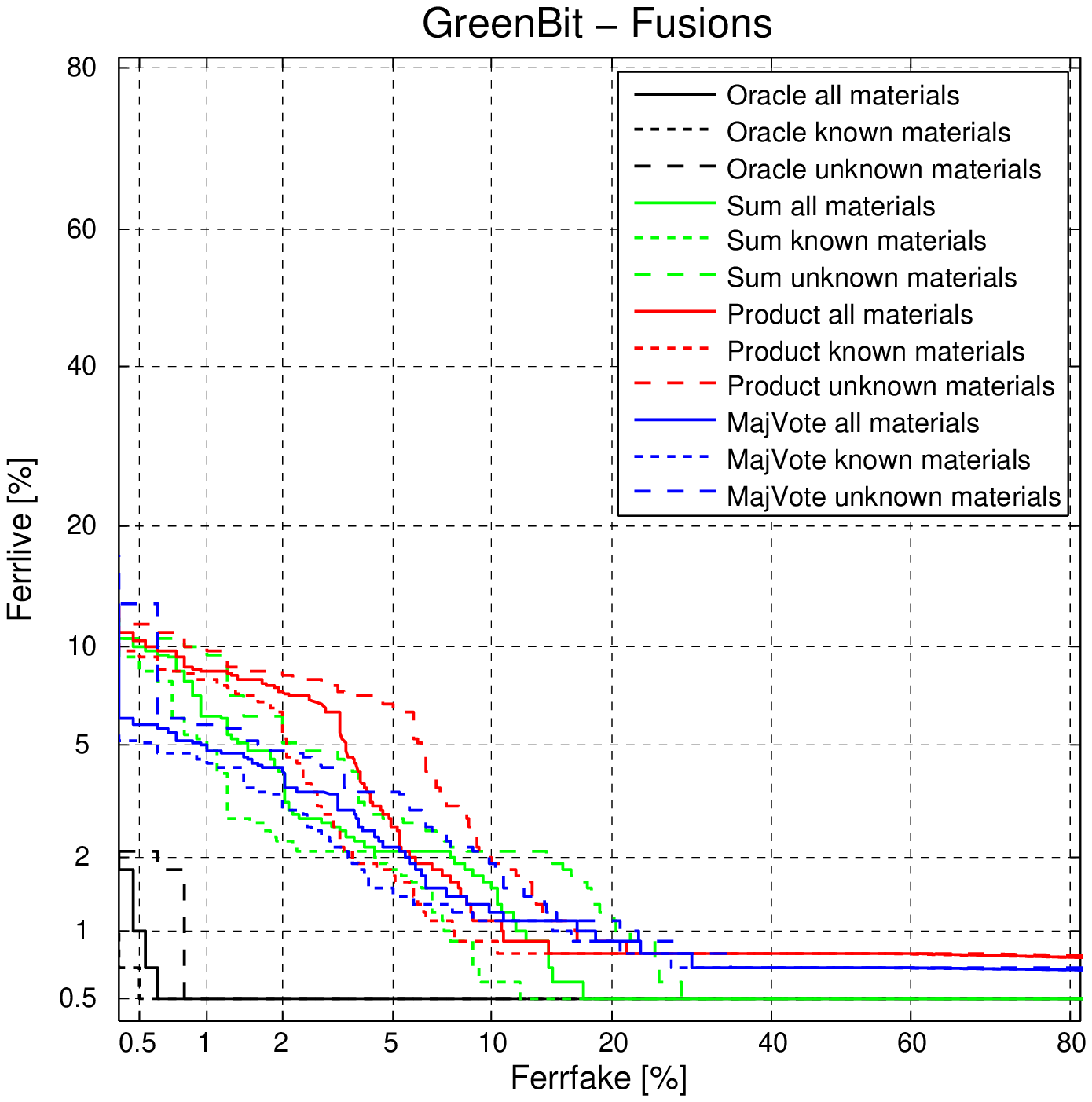}}
\caption{LivDet 2015 Greenbit dataset: DET curves of the nogueira, unina and anonym algorithms (a), DET curves of the fusion of the three classifiers (b).}
\label{fig:rocGreenbit}
\end{figure}

\begin{figure}[t]
\centering
\subfloat[]{\includegraphics[width=0.45\linewidth]{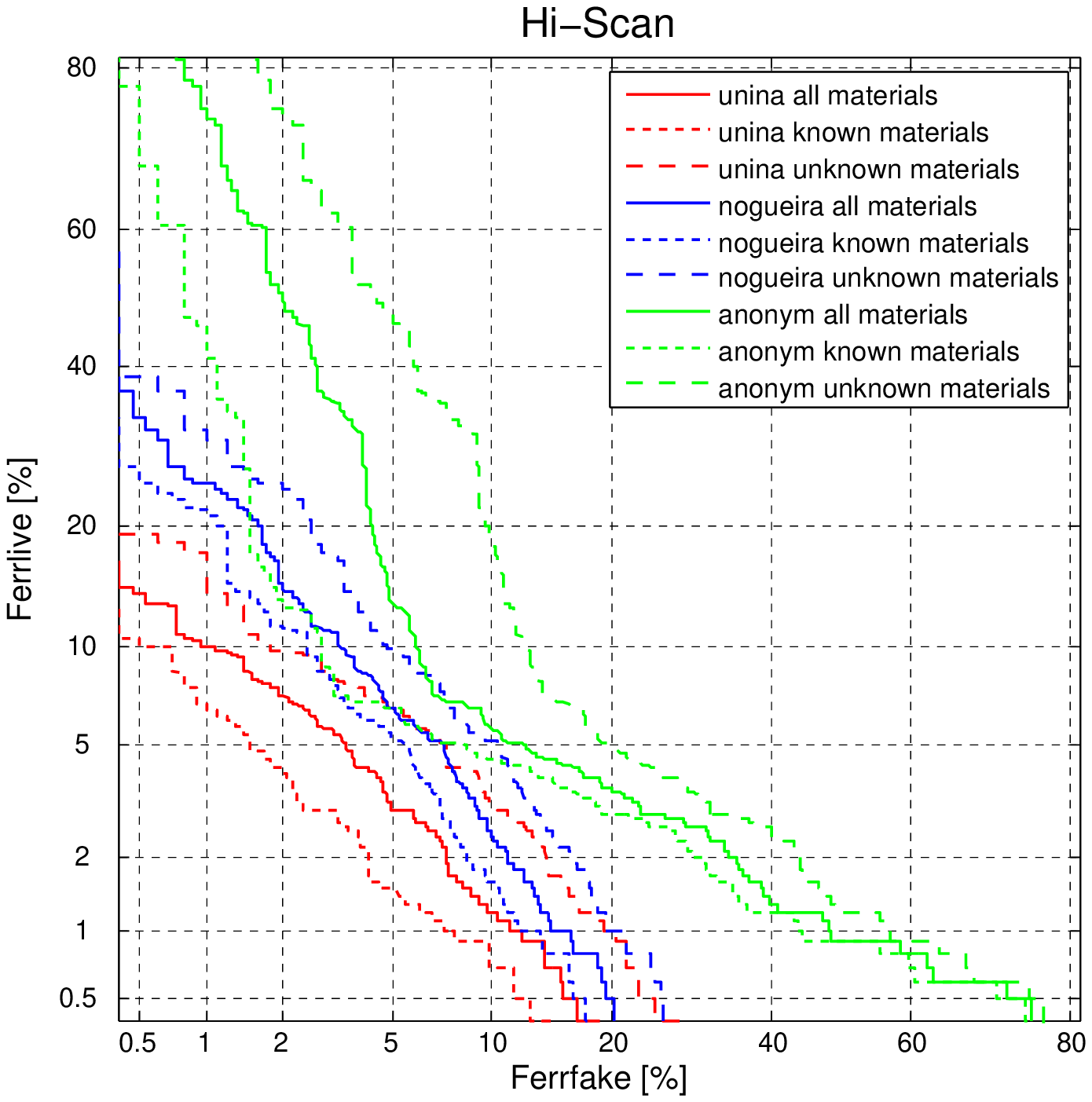}}
\qquad
\subfloat[]{\includegraphics[width=0.45\linewidth]{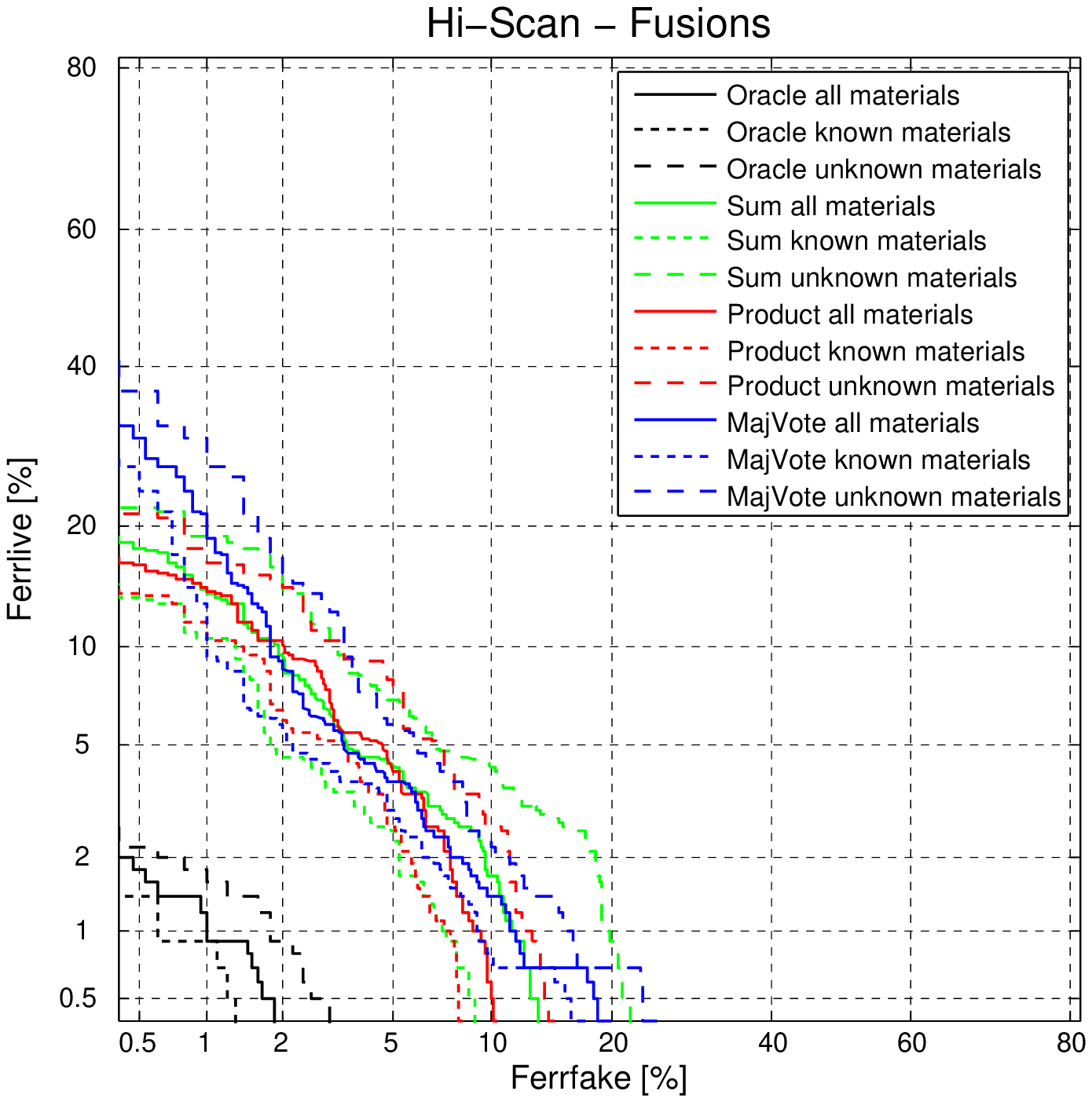}}
\caption{LivDet 2015 Biometrika dataset: DET curves of the nogueira, unina and anonym algorithms (a), DET curves of the fusion of the three classifiers (b).}
\label{fig:rocBiometrika}
\end{figure}

\begin{figure}[t]
\centering
\subfloat[]{\includegraphics[width=0.45\linewidth]{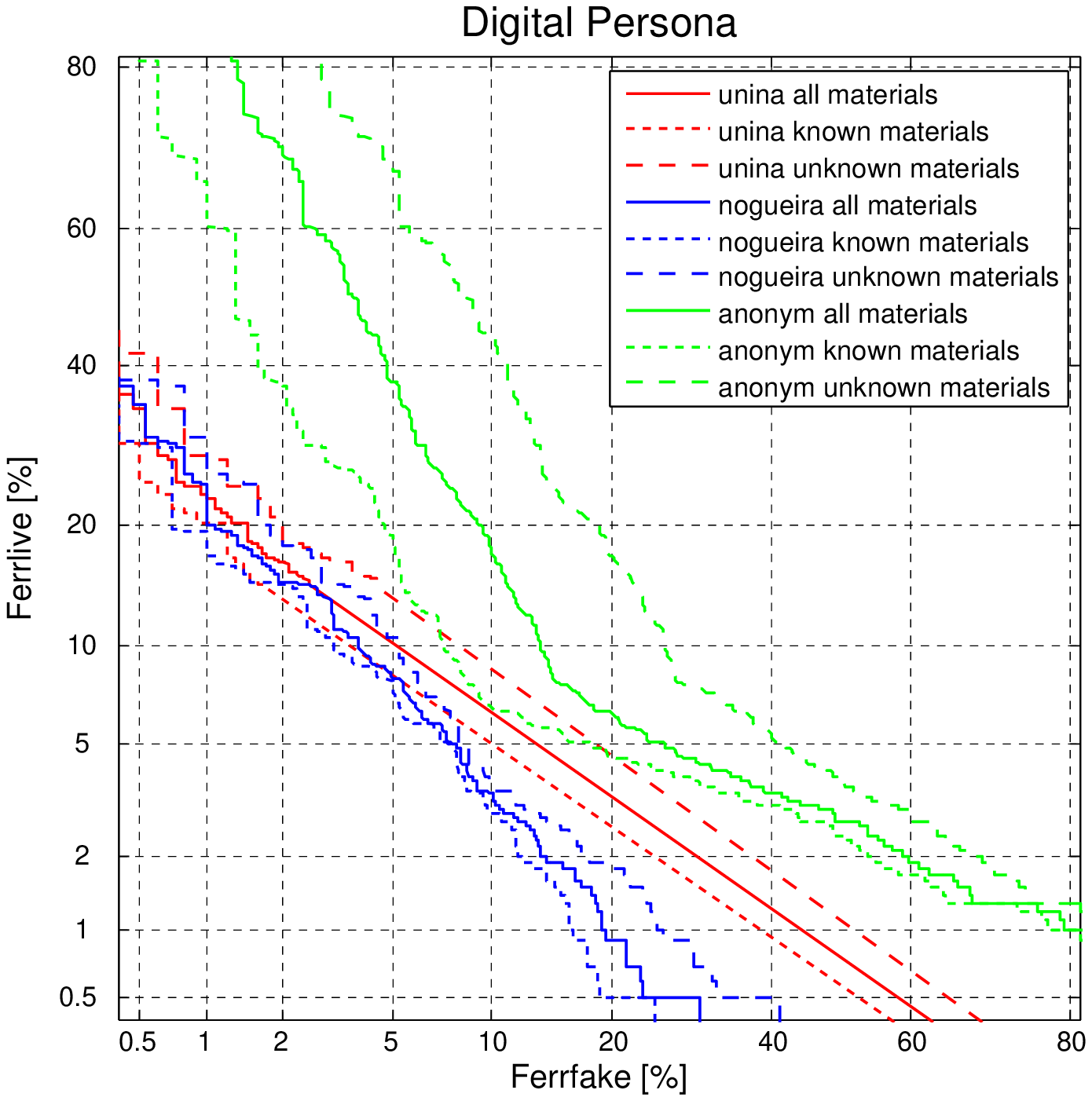}}
\qquad
\subfloat[]{\includegraphics[width=0.45\linewidth]{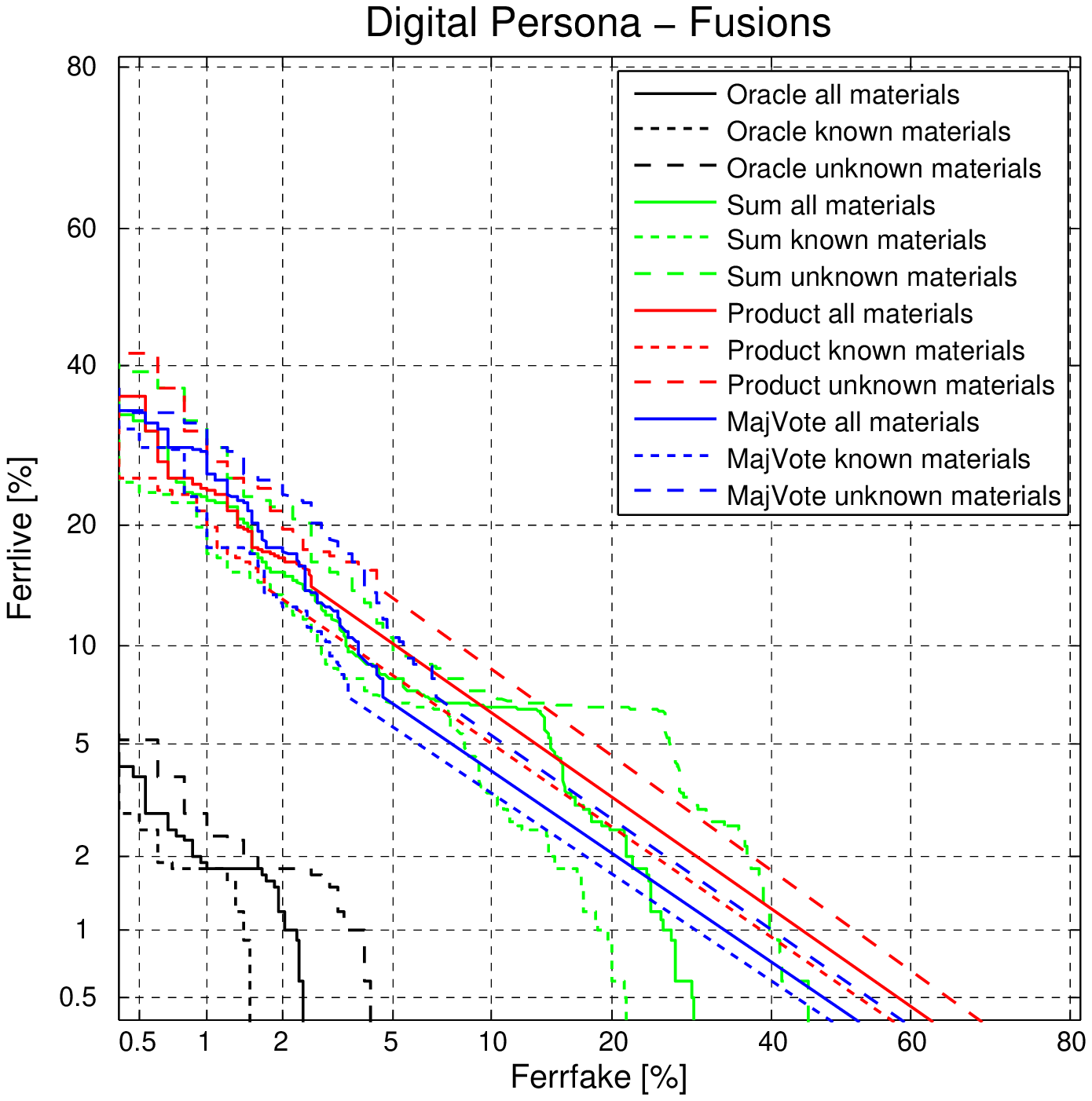}}
\caption{LivDet 2015 Digital Persona dataset: DET curves of the nogueira, unina and anonym algorithms (a), DET curves of the fusion of the three classifiers (b).}
\label{fig:rocDigital}
\end{figure}

\begin{figure}[t]
\centering
\subfloat[]{\includegraphics[width=0.45\linewidth]{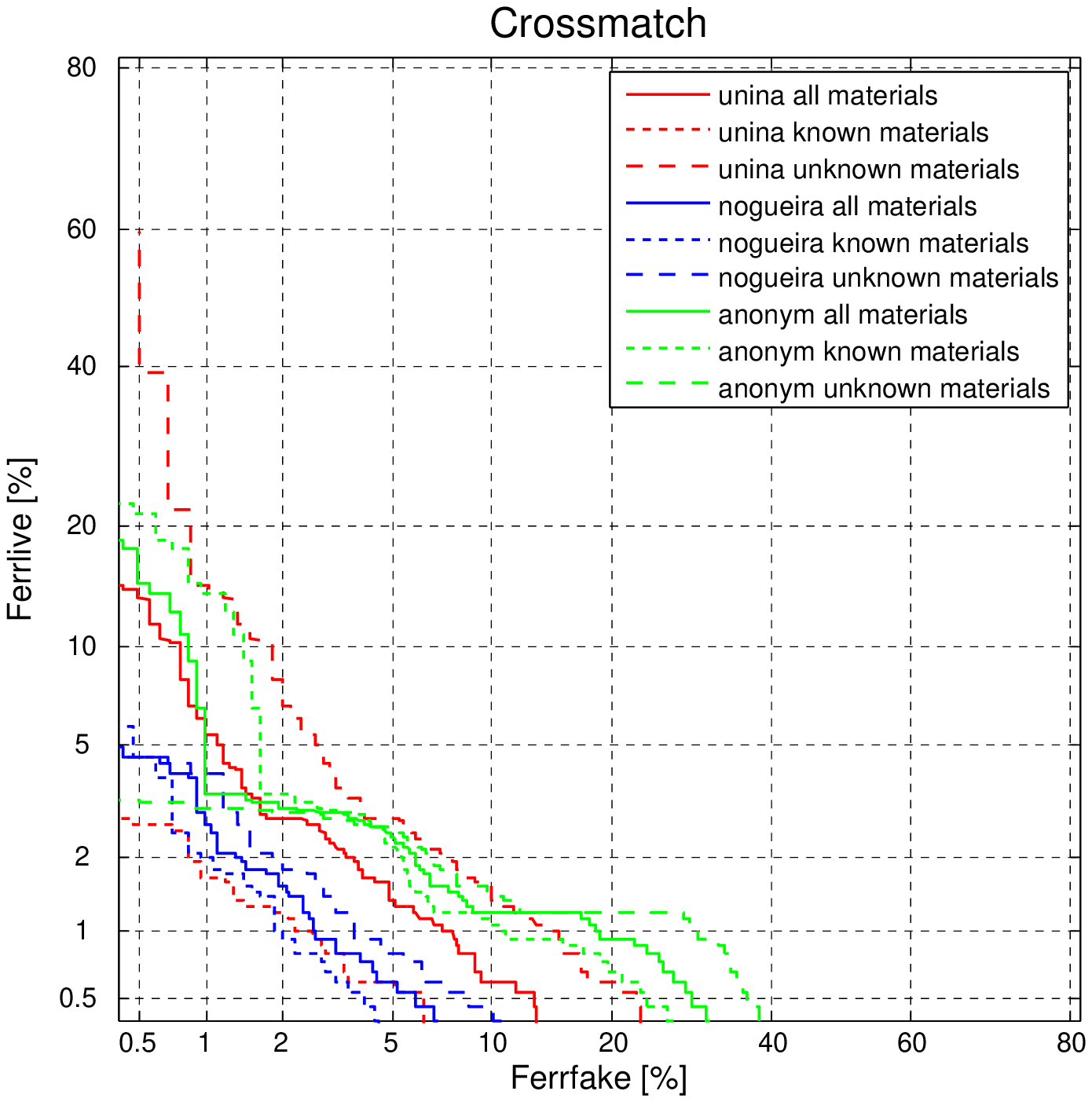}}
\qquad
\subfloat[]{\includegraphics[width=0.45\linewidth]{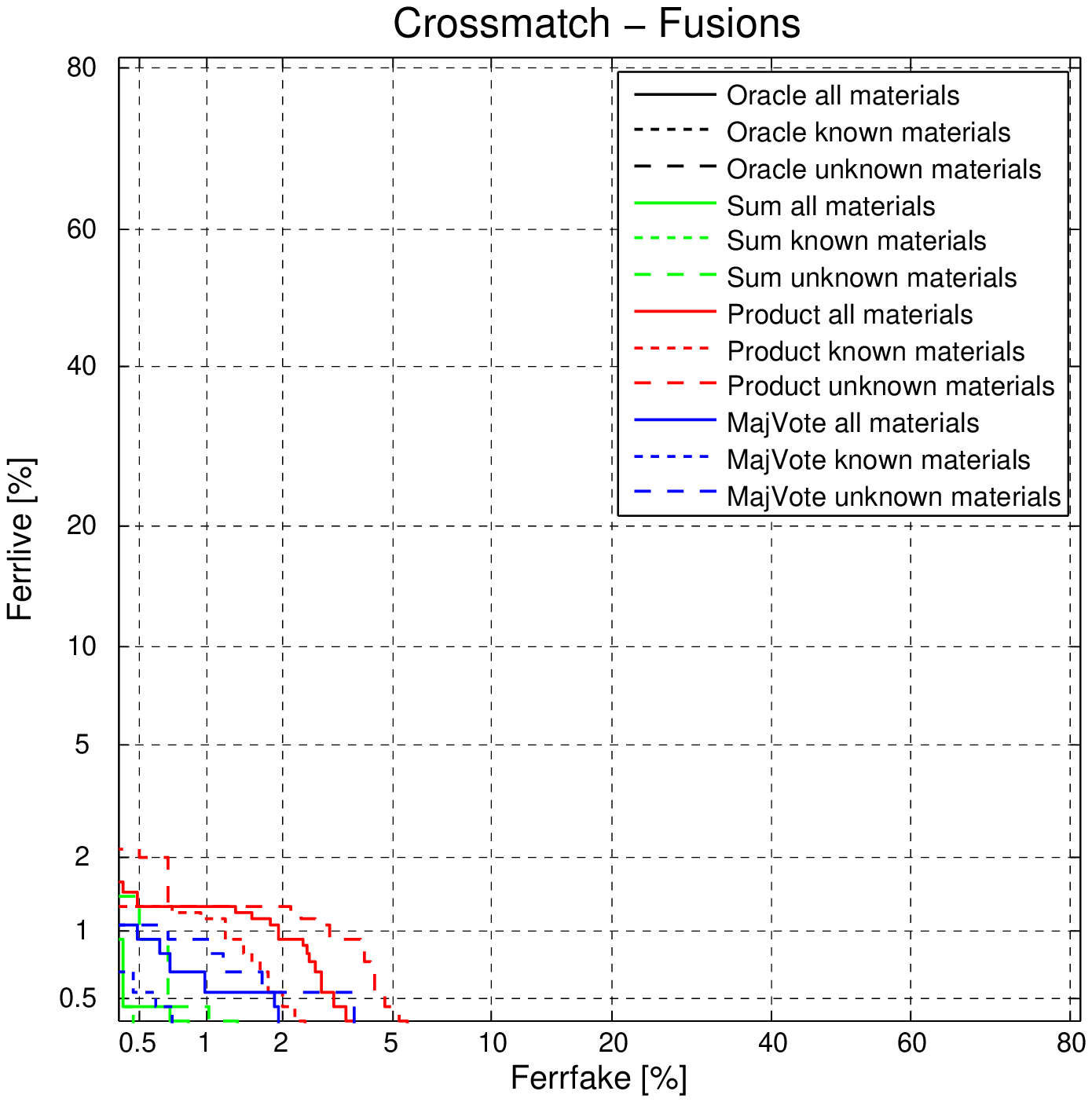}}
\caption{LivDet 2015 Crossmatch dataset: DET curves of the nogueira, unina and anonym algorithms (a), DET curves of the fusion of the three classifiers (b).}
\label{fig:rocCrossmatch}
\end{figure}

The DET (Detection Error Tradeoff) curves in Figures \ref{fig:rocGreenbit} (a), \ref{fig:rocBiometrika} (a), \ref{fig:rocDigital} (a), \ref{fig:rocCrossmatch} (a), show the performance of three of the four best algorithms for LivDet 2015 sorted by ACE (nogueira, unina and anonym). Unfortunately, we could not plot the jinglian's algorithm performance because it output only two possible values, namely 0 or 100. There being no intermediate values it was impossible to obtain different Ferrlive and Ferrfake values varying the threshold value and produce the DET curve. More discussion on fusion results for Figures \ref{fig:rocGreenbit} (b), \ref{fig:rocBiometrika} (b), \ref{fig:rocDigital} (b), \ref{fig:rocCrossmatch} (b) will be provided in further below.

%For each of the three algorithms we plotted three different curves: the one calculated taking into account only the known materials, the one with only the unknown materials and the one with all the materials. Obviously, for each algorithm, the known materials curve always perform better than the unknown materials.

Another important indicator of an algorithm validity is the Ferrfake value calculated when $Ferrlive=1\%$. This value represent the percentage of spoofs able to hack into the system when the rate of legitimate users that are rejected is no more than 1\%.
As a matter of fact, by varying the threshold value, different Ferrfake and Ferrlive values are obtained and, as the threshold grows from 0 to 100, Ferrfake decrease and Ferrlive increase.
Obviously ferrlive value must be kept low to minimize the inconvenience to authorized users but, just as important, the low ferrfake value limits the number of unauthorized users able to enter into the system.

Results in Tables \ref{tab:Ferrlive1CMT}, \ref{tab:Ferrlive1DP}, \ref{tab:Ferrlive1GB} and \ref{tab:Ferrlive1BMK} show that even the best performing algorithm (nogueira) is not yet good enough since when $Ferrlive=1\%$, the ferrfake values (testing on all materials) ranges from 2.66\% to 19.10\%. These are the percentage  of unauthorized users that the system is unable to correctly classify. If we consider only the unknown materials, the results are even worse, ranging from 3.69\% to 25.30\%.

On the basis of such results, we can say that there is no specific algorithm, among the analyzed ones, able to generalize against never-seen-before spoofing attacks. We observed a performance drop and also found that the amount of the drop is unpredictable as it depends on the material. This should be matter of future discussions.

\begin{table*}[p]
\begin{center}
\begin{tabular}[t]{ | l || c | c | c |}\hline
	& \textbf{all materials} & \textbf{known materials}	& \textbf{unknown materials}\\ \hline
unina	&	7.42	&	2.47	&	14.49	\\
nogueira	&	2.66	&	1.94	&	3.69	\\
anonym	&	18.61	&	10.75	&	29.82	\\ \hline
average	&	9.56	&	5.05	&	16.00	\\
std. dev.	&	8.19	&	4.94	&	13.13	\\ \hline
\end{tabular}
\caption{Ferrfake values of the best algorithms calculated when $Ferrlive=1\%$ for the Crossmatch dataset.}
\label{tab:Ferrlive1CMT}
\end{center}
\end{table*}

\begin{table*}[p]
\begin{center}
\begin{tabular}[t]{ | l || c | c | c |}\hline
	& \textbf{all materials} & \textbf{known materials}	& \textbf{unknown materials}\\ \hline
unina	&	51.30	&	50.85	&	52.20	\\
nogueira	&	19.10	&	16.00	&	25.30	\\
anonym	&	80.83	&	79.25	&	84.00	\\ \hline
average	&	50.41	&	48.70	&	53.83	\\
std. dev.	&	30.87	&	31.68	&	29.38	\\ \hline
\end{tabular}
\caption{Ferrfake values of the best algorithms calculated when $Ferrlive=1\%$ for the Digital Persona dataset.}
\label{tab:Ferrlive1DP}
\end{center}
\end{table*}

\begin{table*}[p]
\begin{center}
\begin{tabular}[t]{ | l || c | c | c |}\hline
	& \textbf{all materials} & \textbf{known materials}	& \textbf{unknown materials}\\ \hline
unina	&	41.80	&	36.10	&	53.20	\\
nogueira	&	17.90	&	15.15	&	23.40	\\
anonym	&	75.47	&	75.25	&	75.90	\\ \hline
average	&	45.06	&	42.17	&	50.83	\\
std. dev.	&	28.92	&	30.51	&	26.33	\\ \hline
\end{tabular}
\caption{Ferrfake values of the best algorithms calculated when $Ferrlive=1\%$ for the Green Bit dataset.}
\label{tab:Ferrlive1GB}
\end{center}
\end{table*}

\begin{table*}[p]
\begin{center}
\begin{tabular}[t]{ | l || c | c | c |}\hline
	& \textbf{all materials} & \textbf{known materials}	& \textbf{unknown materials}\\ \hline
unina	&	11.60	&	7.50	&	19.80	\\
nogueira	&	15.20	&	12.60	&	20.40	\\
anonym	&	48.40	&	44.05	&	57.10	\\ \hline
average	&	25.07	&	21.38	&	32.43	\\
std. dev.	&	20.29	&	19.79	&	21.36	\\ \hline
\end{tabular}
\caption{Ferrfake values of the best algorithms calculated when $Ferrlive=1\%$ for the Biometrika dataset.}
\label{tab:Ferrlive1BMK}
\end{center}
\end{table*}
\clearpage

\begin{figure}[p]
\centering 
\subfloat[]{\includegraphics[width=0.45\linewidth]{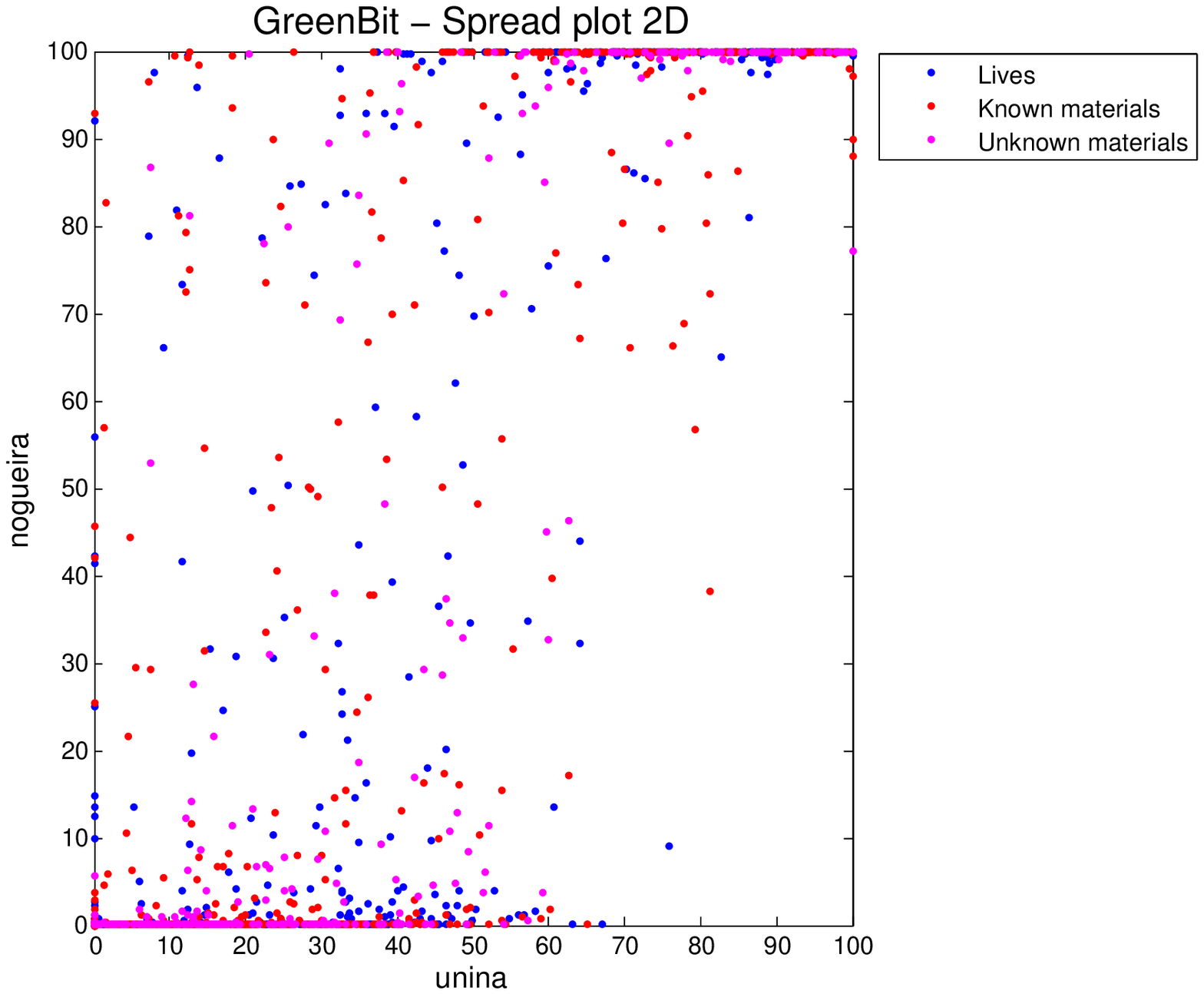}}
\qquad
\subfloat[]{\includegraphics[width=0.45\linewidth]{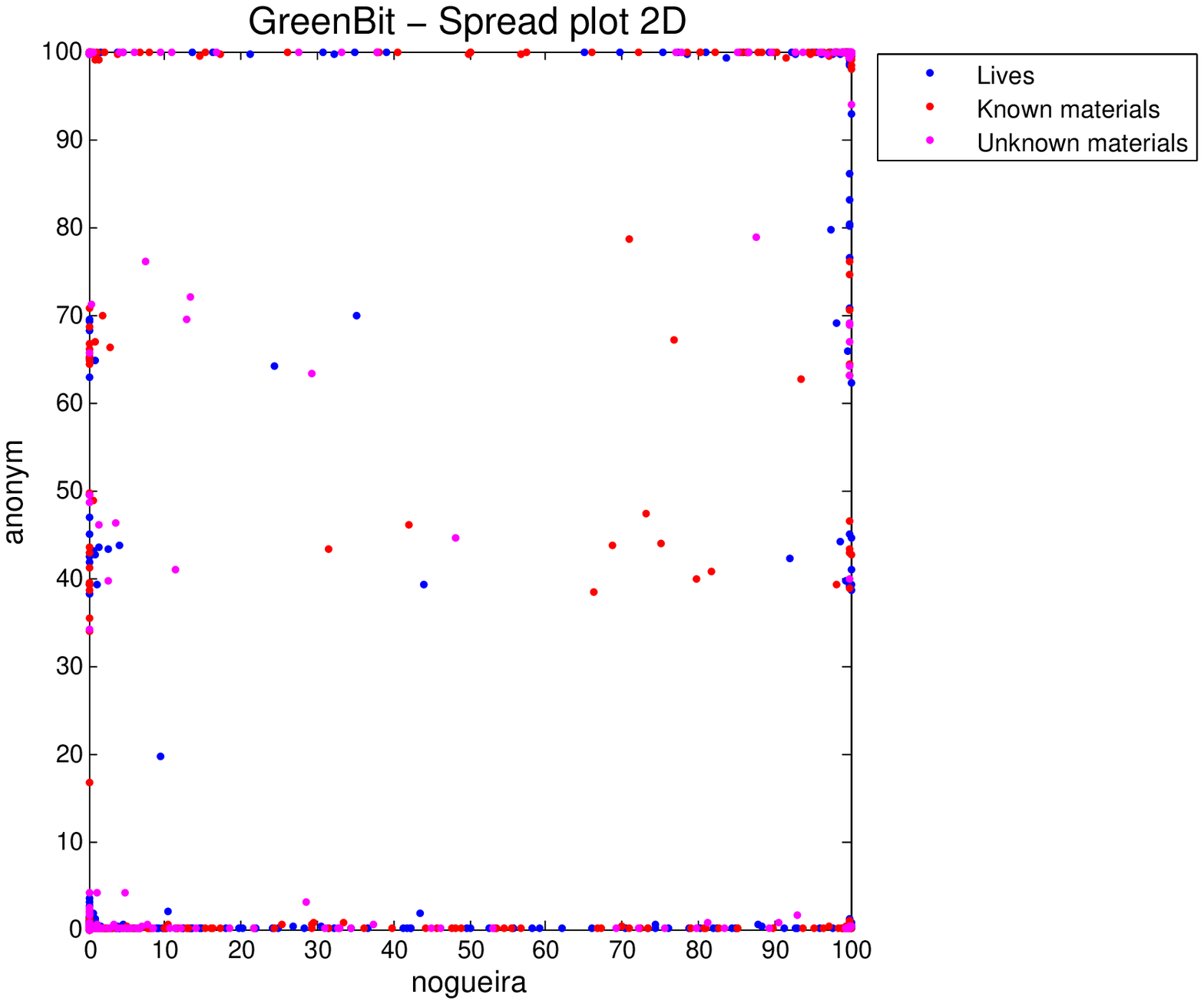}}\\
%\qquad
\subfloat[]{\includegraphics[width=0.45\linewidth]{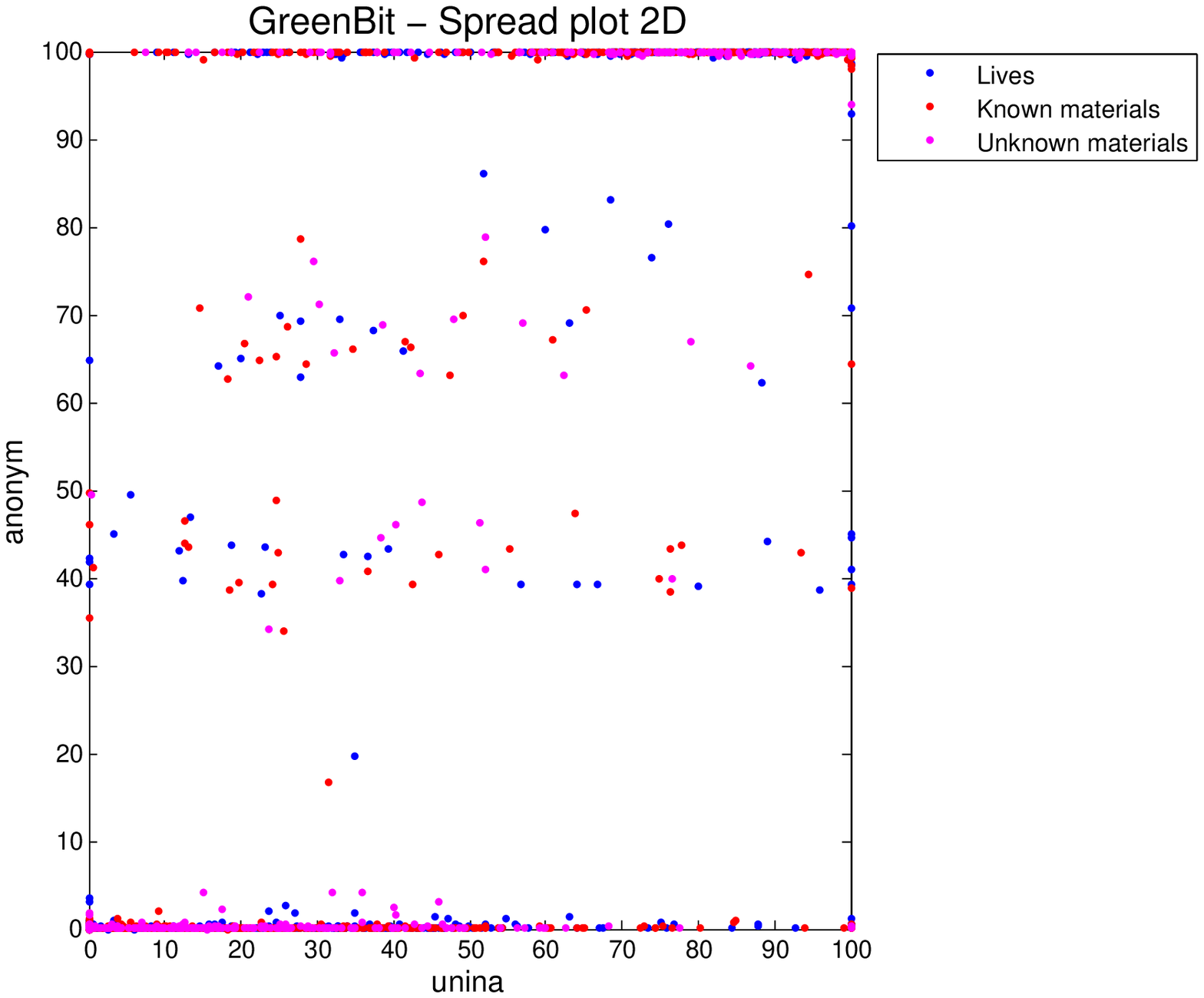}}
\qquad
\subfloat[]{\includegraphics[width=0.45\linewidth]{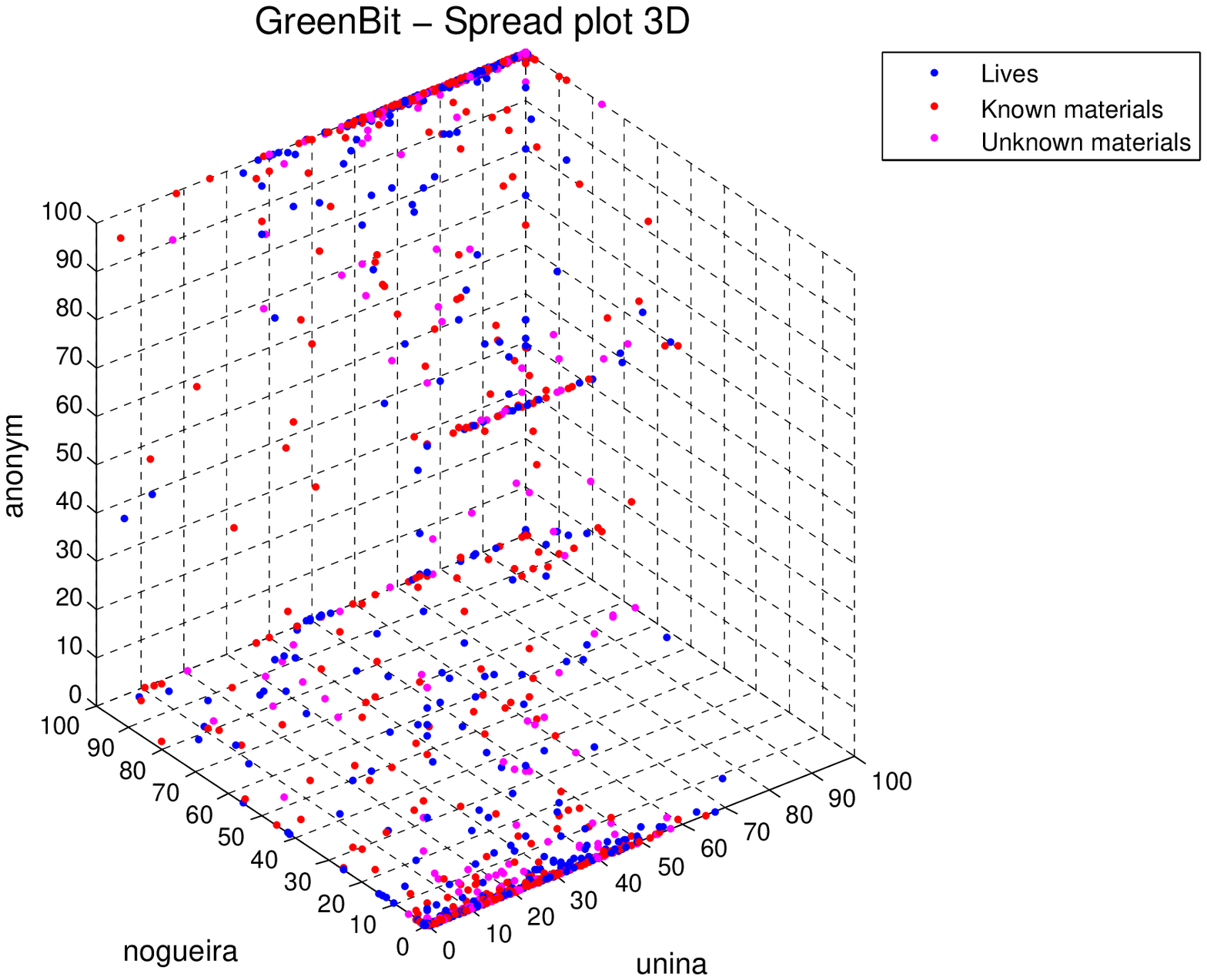}}
\caption{2D and 3D spread plots for the Greenbit dataset: (a) unina - nogueira, (b) nogueira - anonym, (c) unina - anonym, (d) nogueira - unina - anonym.}
\label{fig:greenbitPlot}
\end{figure}

\begin{figure}[p]
\centering
\subfloat[]{\includegraphics[width=0.45\linewidth]{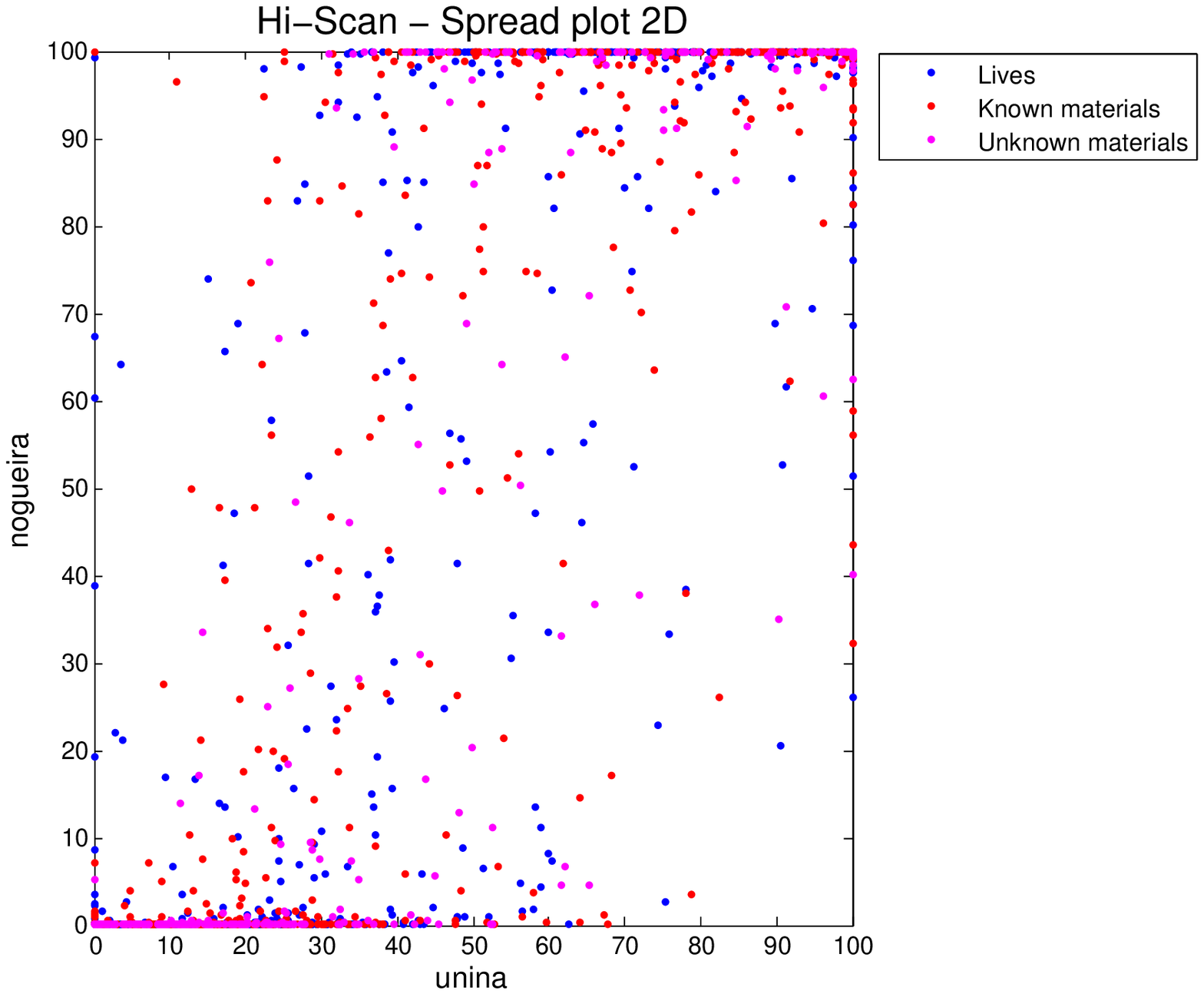}}
\qquad
\subfloat[]{\includegraphics[width=0.45\linewidth]{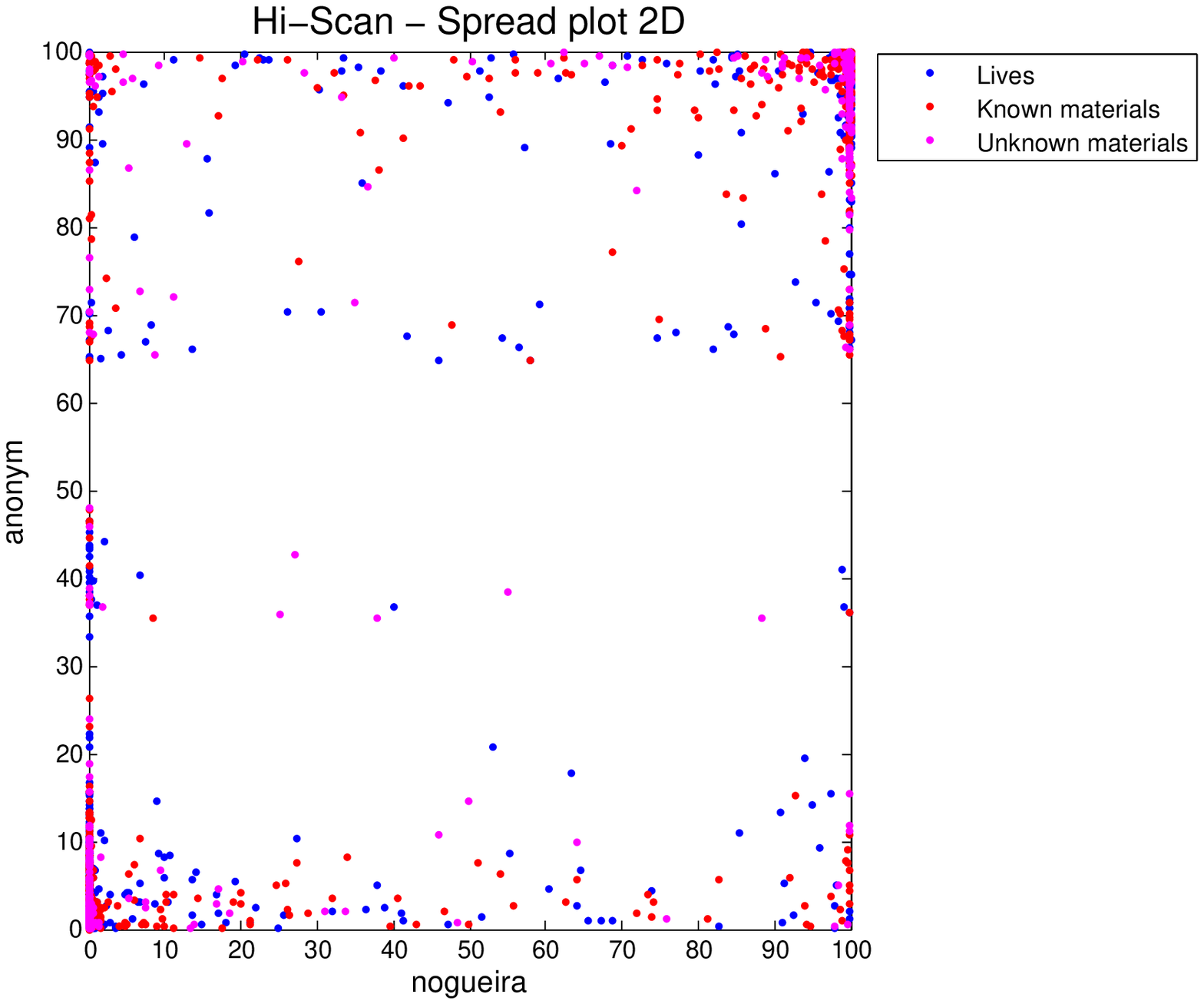}}\\
%\qquad
\subfloat[]{\includegraphics[width=0.45\linewidth]{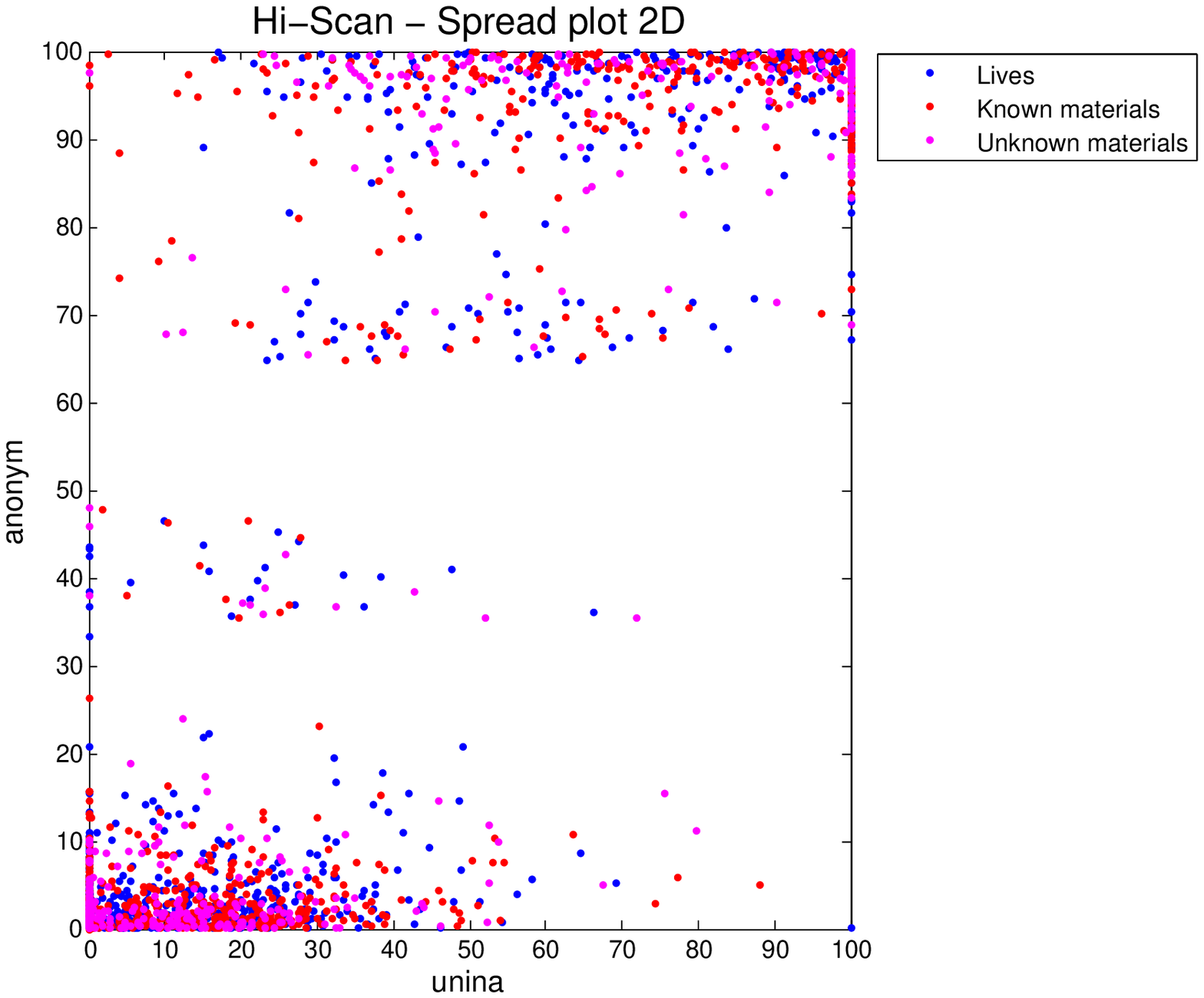}}
\qquad
\subfloat[]{\includegraphics[width=0.45\linewidth]{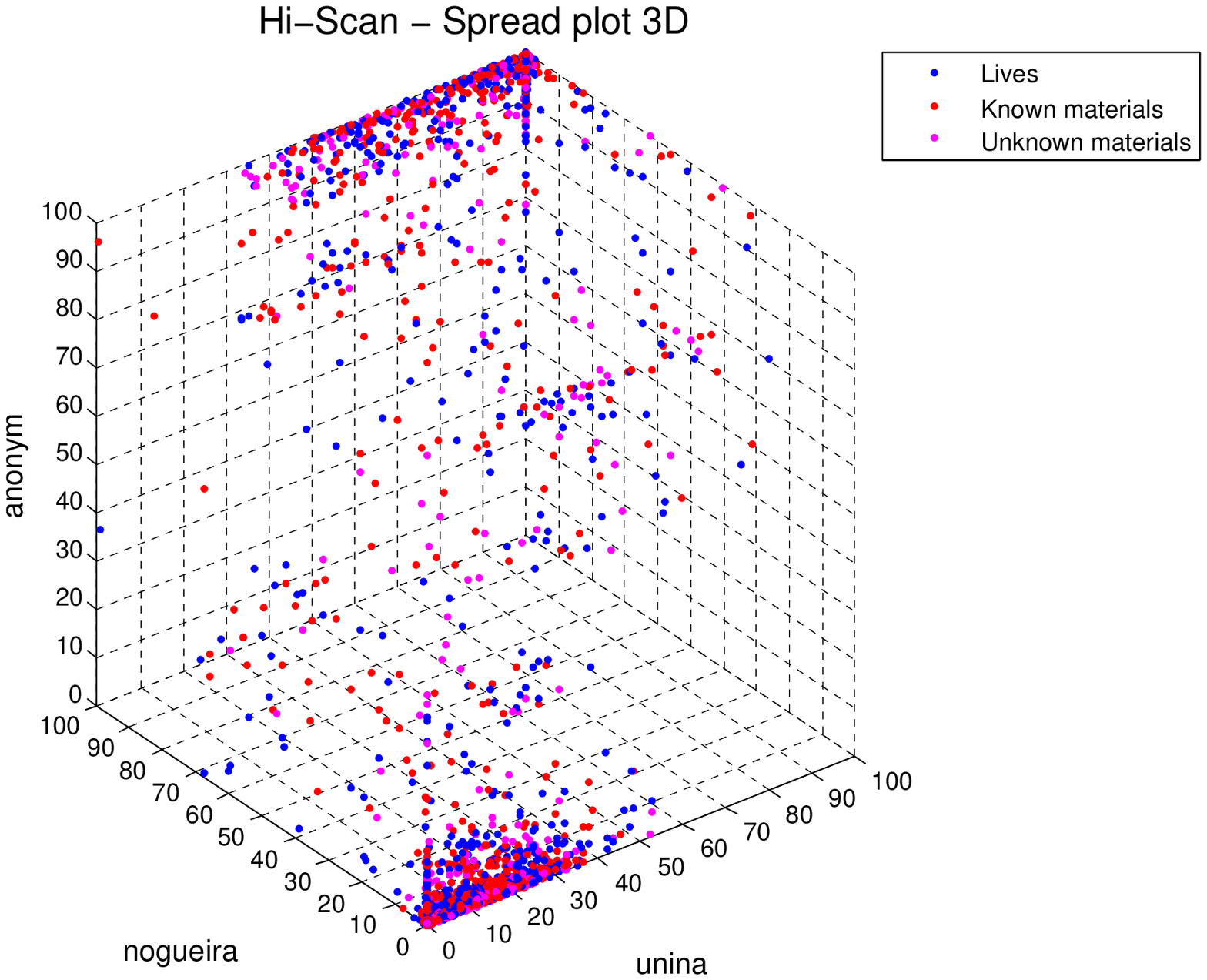}}
\caption{2D and 3D spread plots for the Biometrika dataset: (a) unina - nogueira, (b) nogueira - anonym, (c) unina - anonym, (d) nogueira - unina - anonym.}
\label{fig:hiscanPlot}
\end{figure}

\begin{figure}[p]
\centering
	\subfloat[]{\includegraphics[width=0.45\linewidth]{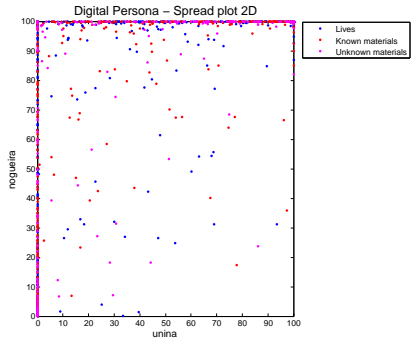}}
\qquad
\subfloat[]{\includegraphics[width=0.45\linewidth]{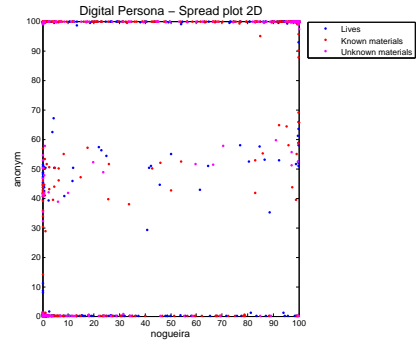}}\\
%\qquad
\subfloat[]{\includegraphics[width=0.45\linewidth]{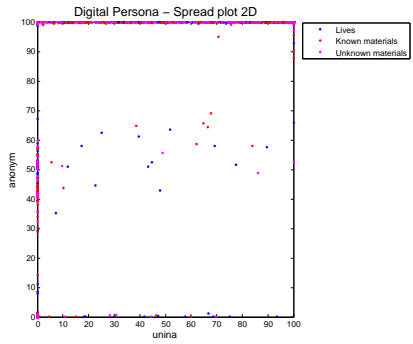}}
\qquad
\subfloat[]{\includegraphics[width=0.45\linewidth]{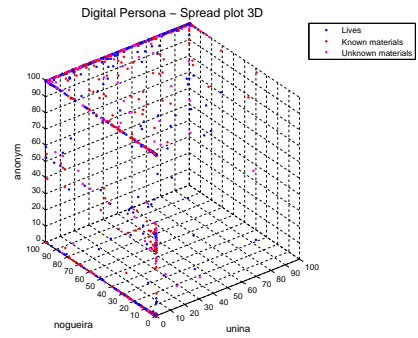}}
\caption{2D and 3D spread plots for the Digital Persona dataset: (a) unina - nogueira, (b) nogueira - anonym, (c) unina - anonym, (d) nogueira - unina - anonym.}
\label{fig:digitalPlot}
\end{figure}

\begin{figure}[p]
\centering
\subfloat[]{\includegraphics[width=0.45\linewidth]{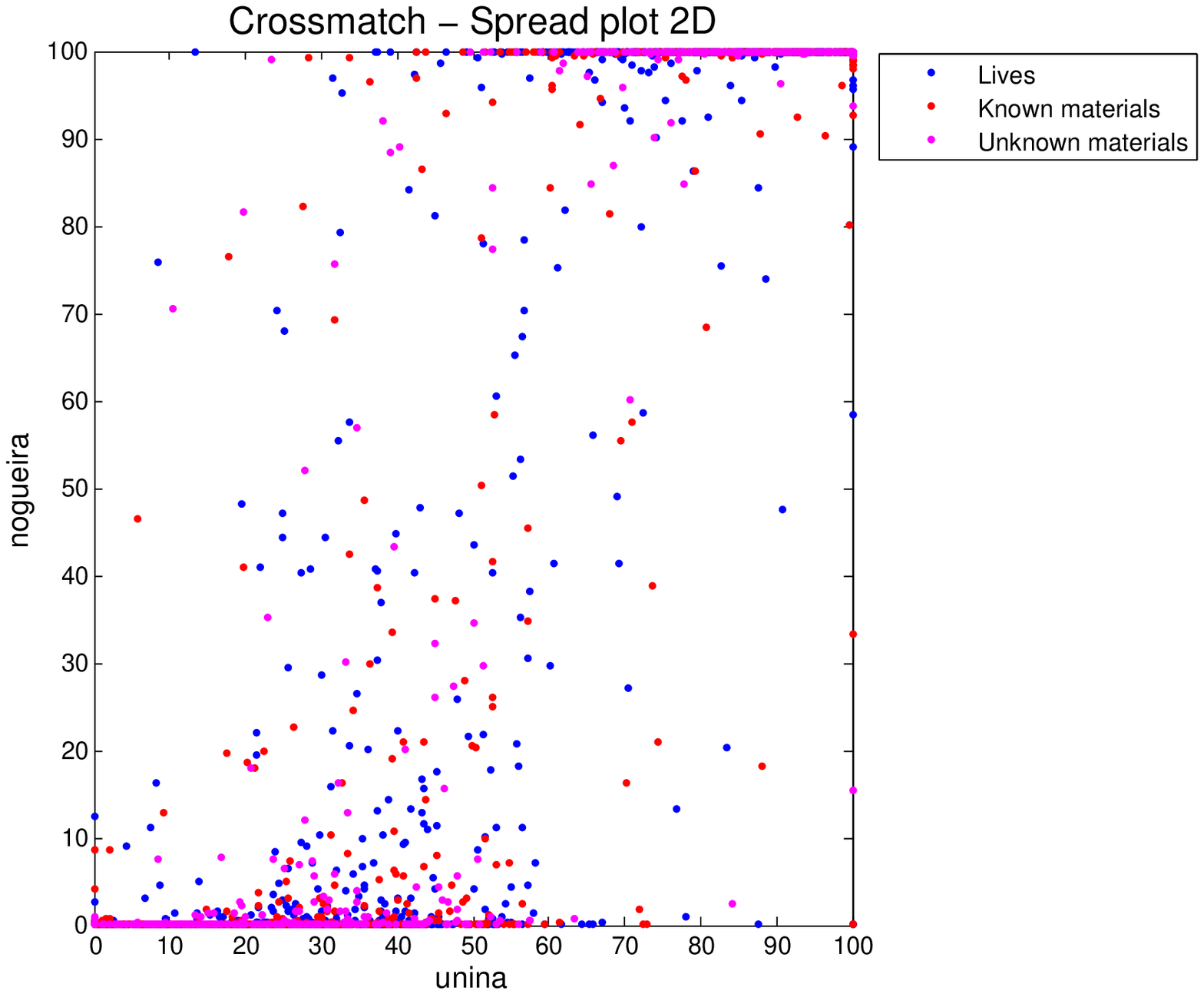}}
\qquad
\subfloat[]{\includegraphics[width=0.45\linewidth]{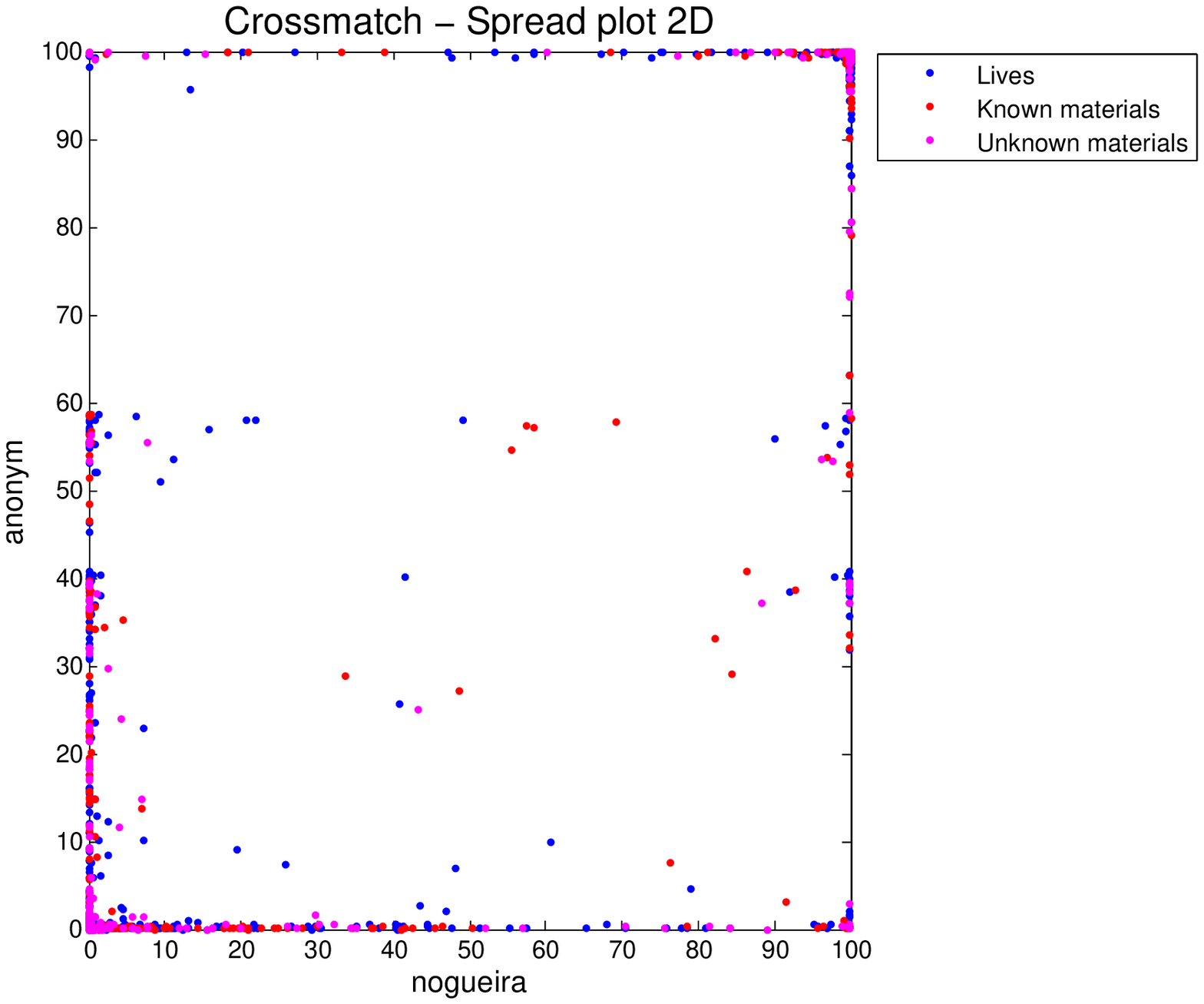}}\\
%\qquad
\subfloat[]{\includegraphics[width=0.45\linewidth]{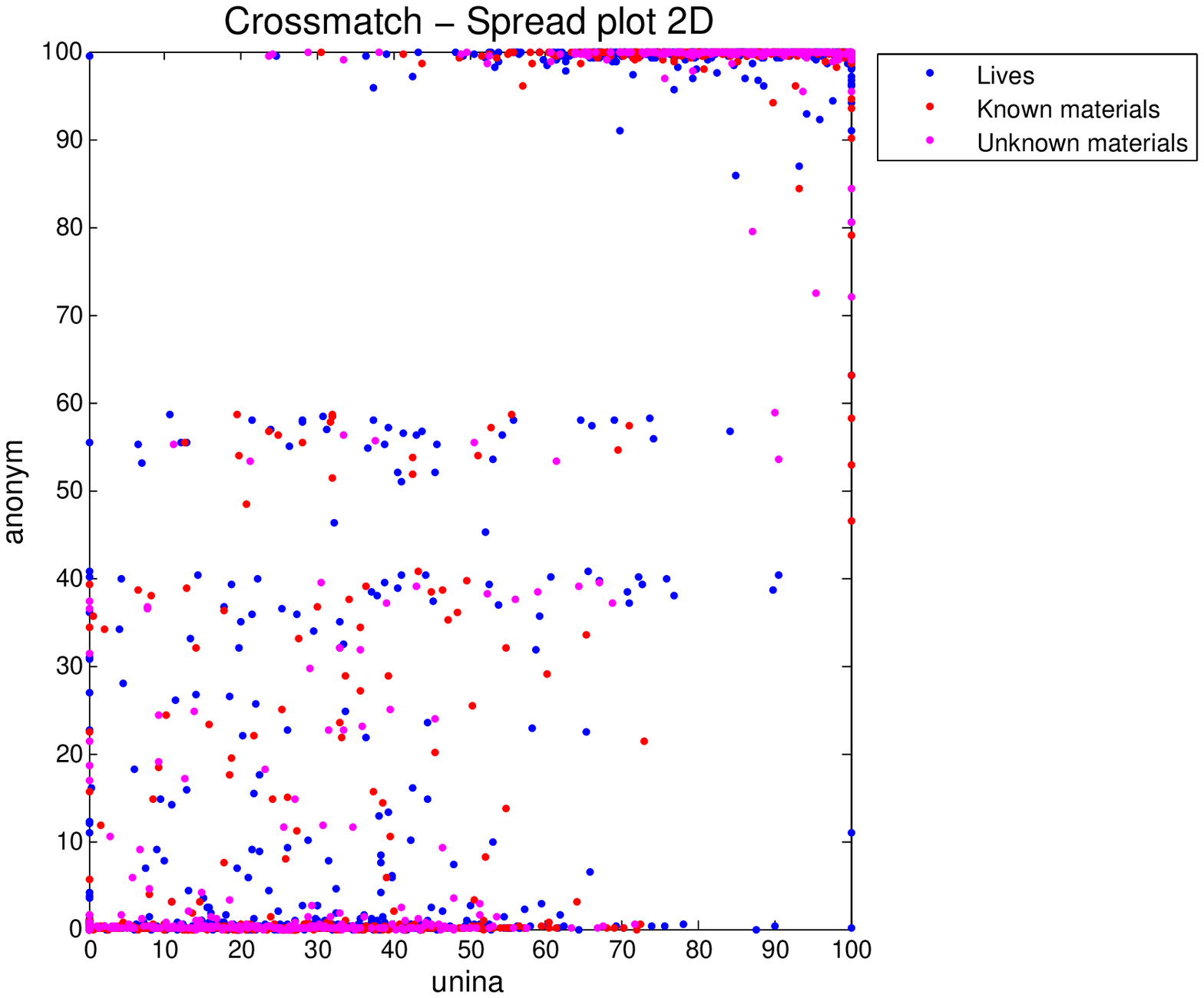}}
\qquad
\subfloat[]{\includegraphics[width=0.45\linewidth]{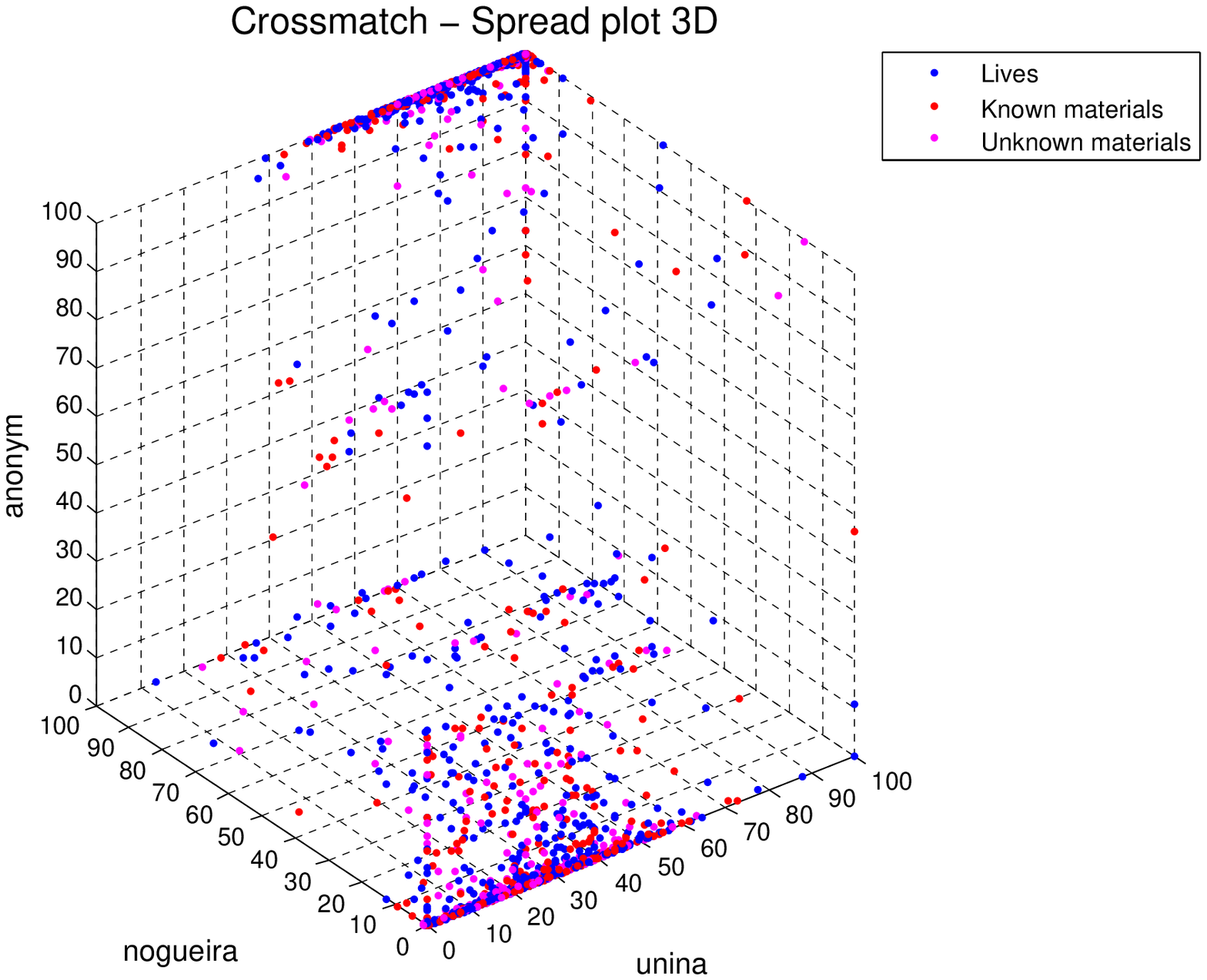}}
\caption{2D and 3D spread plots for the Crossmatch dataset: (a) unina - nogueira, (b) nogueira - anonym, (c) unina - anonym, (d) nogueira - unina - anonym.}
\label{fig:crossmatchPlot}
\end{figure}

On the other hand, the performance difference among algorithms may be complementary, that is, one algorithm may be able to detect spoofs that another one does not. Accordingly, Figures \ref{fig:greenbitPlot}, \ref{fig:hiscanPlot}, \ref{fig:digitalPlot} and \ref{fig:crossmatchPlot} show, for each dataset, the 2D and 3D spread plots generated using again three out of the four better algorithms. In the 3D plots each point coordinate corresponds to the three match scores obtained by the three algorithms for each image. In the 2D plots the same results are presented using two algorithms at a time for a total of three 2D plots for each 3D plot.
These plots give an idea of the correlation level of the spoof detection ability of the selected algorithms.
As a matter of fact, it appears that in many doubtful cases, the uncertainty of a classifier does not correspond to that of the others. For cases which have a high correlation, the points in the graph would be more or less along a diagonal line. However on this plot, the values are well distributed and this indicate a lower correlation between the algorithms and thus may be complementary.

Since there is evidence that the submitted algorithms may be complementary, we performed three fusion techniques at the score level using again three out of the four better algorithms. We used the sum of the \textit{a posteriori} probabilities generated by the three algorithms, the product of the same \textit{a posteriori} probabilities and the majority vote (each fingerprint image is classified as live or fake if the majority of individual algorithms rank it as live or fake).
These three fusion techniques have very different characteristics from each other: the product tends to crush on a negative classification (i.e. the pattern rejection) rather than positive (for the zero-product property); the sum instead has an opposite characteristic namely it smooths the differences (such as a low-pass filter); finally, the majority emphasize where classifiers agree.
%Since the outputs produced by the third algorithm were only two values, 0 or 100 we use his values for the majority vote but we had to use the fourth algorithm for the sum and product rules.
In order to show the pros and cons of these classifiers fusion we also plot the curve of the so called ``Oracle response". The Oracle is an ideal fusion technique that correctly classifies a pattern if that pattern is correctly classified by at least one of the single classifiers. It indicates the maximum accuracy that can be potentially achieved.
Obviously, the more the results of the various classifiers are complementary the better are the results of the oracle.

The DET curves in Figures \ref{fig:rocGreenbit} (b), \ref{fig:rocBiometrika} (b), \ref{fig:rocDigital} (b), \ref{fig:rocCrossmatch} (b), compared with that of the single classifiers, clearly show a result improvement especially when using the sum rule. As a matter of fact, as stated before, sum smooths the differences. If score values are balanced (that is x for accept and 1-x for reject) the sum rule emphasizes where classifiers agree as for the majority vote. Moreover, since it averages the results, little doubts of two (wrong) classifiers may be counterbalanced by a strong performance in the (correct) third, leading to better results.
The only case in which the performance declines is that of the Digital Persona dataset.
Although the fusion at score-level appear a good solution to improve the performance, current results suggest that individual algorithms must be carefully selected and also appropriate combination methodologies should be studied as well.

\subsection{Trends of Competitors and Results for Fingerprint Part 2: Systems}
\label{sec:trends2}

Fingerprint Part 2: Systems has not yet shown growth in the number of competitors, but it has only existed for 3 competitions thus far. There are more limitations to the systems testing portion.  It is not surprising that there are not great numbers of entrants given the need for a full fingerprint recognition device to be built with an integrated liveness detection module. There has also been a general lack of interest in companies shipping full systems for testing and it appears that there is more comfort in submitting an algorithm. Both Livdet 2011 and LivDet 2013 had 2 submissions whilst LivDet 2015 had only one. Information about competitors is shown in Table \ref{tab:participants2}.

\begin{table*}[!htb]
\begin{center}
\begin{tabular}[t]{ | c | c |}
\hline
\textbf{Participants LivDet 2011} & \textbf{Algorithm Name} \\ \hline\hline
Dermalog Identification Systems GmbH & Dermalog\\ \hline
GreenBit & GreenBit\\ \hline\hline

\textbf{Participants LivDet 2013} & \textbf{Algorithm Name} \\ \hline\hline
Dermalog Identification Systems GmbH & Dermalog\\ \hline
Anonymous & Anonymous1\\ \hline\hline

\textbf{Participants LivDet 2015} & \textbf{Algorithm Name} \\ \hline\hline
Anonymous & Anonymous2\\ \hline

\end{tabular}
\caption{Participants for Part 2: Systems.}
\label{tab:participants2}
\end{center}
\end{table*}

This portion of the LivDet competitions has distinct recognition for the rapid decrease in error rates. In the span of 2 years, the best results from LivDet 2011 were worse than the worst results of LivDet 2013.  Thus, systems tests showed a quicker decrease in error rates as well as the one systems submission in 2013 had lower error rates than any submitted algorithms in LivDet.

In 2011, Dermalog performed at a FerrLive of 42.5\% and a FerrFake of 0.8\%. GreenBit performed at a FerrLive of 38.8\% and a FerrFake of 39.47\%. Both systems had high FerrLive scores.

The 2013 edition produced much better results since Dermalog performed at a FerrLive of 11.8\% and a FerrFake of 0.6\%. Anonymous1 performed at a FerrLive of 1.4\% and a FerrFake of 0.0\%. Both systems had low FerrFake rates. Anonymous1 received a perfect score of 0.0\% error, successfully determining every spoof finger presented as a spoof.

Anonymous2, in 2015, scored a FerrLive of 14.95\% and a FerrFake of 6.29\% at the (given) threshold of 50 \ref{tab:x} showing an improvement over the general results seen in LivDet 2011, however the anonymous system did not perform as well as what was seen in LivDet 2013. There is an 11.09\% FerrFake for known recipes and 1\% for unknown recipes (\ref{tab:x}). This result is opposite what has been seen in previous LivDet competitions where known spoof types typically have a better performance than unknown spoof types. The error rate for spoof materials was primarily due to impact on color differences error for the playdoh. Testing across 6 different colors of playdoh found that certain colors behaved in different ways. For yellow and white playdoh, the system detected spoofs as fake with high accuracy.  For brown and black playdoh, the system would not collect an image. Therefore, it was recorded as a fake non-response and not an error in detection of spoofs. For pink and lime green playdoh, the system incorrectly accepted spoofs as live for almost 100\% of images collected. The fact that almost all pink and lime green playdoh images were accepted as live images resulted in a 28\% total error rate for playdoh. The system had a 6.9\% Fake Non-Response Rate primarily due to brown and black playdoh.  This is the first LivDet competition where color of playdoh has been examined in terms of error rates. 

\begin{figure}[t]
\begin{center}
  \includegraphics[width=1\linewidth]{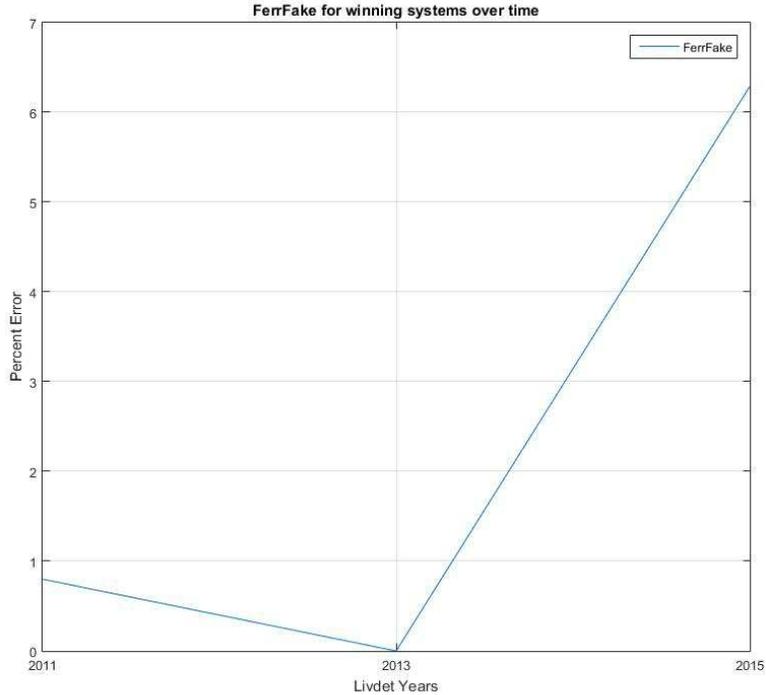}
\end{center}
\caption{FerrFake for winning systems by year .}
\label{fig:FFbyyear}       % Give a unique label
\end{figure}

\begin{figure}[t]
\begin{center}
  \includegraphics[width=1\linewidth]{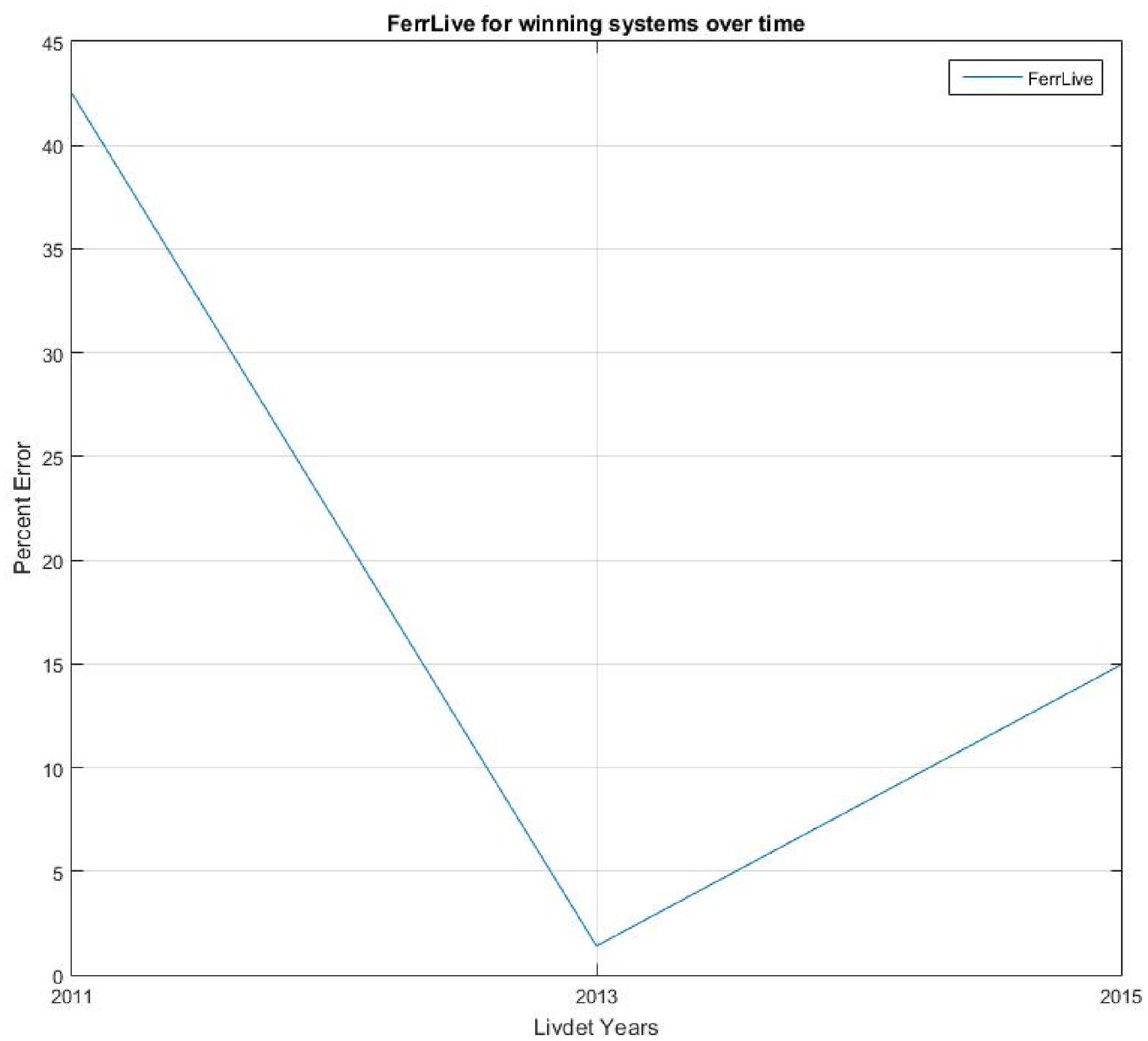}
\end{center}
\caption{FerrLive for winning systems by year .}
\label{fig:FLbyyear}       % Give a unique label
\end{figure}

Examining the trends of results over the three competitions has shown that since 2011 there has been a downward trend in error rates for the systems. Ferrlive in 2015 while higher than 2013, is drastically lower than 2011. FerrFake has had similar error rates over the years. While the 2015 competition showed a 6.29\% Ferrfake, the majority of that error stems from playdoh materials, particularly pink and lime colors. If you discount the errors seen in playdoh, the FerrFake is below 2\% for the 2015 system. The error rates for winning systems over the years is shown in Figure \ref{fig:FFbyyear} and Figure \ref{fig:FLbyyear}

\begin{table}[p]
\begin{center}
\begin{tabular}[t]{ | l || c | c|}
\hline
Submitted System & FerrLive & FerrFake \\ \hline
Anonymous & 14.95\% & 6.29\%\\  \hline\hline
Submitted System & FerrFake  & FerrFake  \\
 &  Known &  Unknown \\ \hline
Anonymous & 11.09\% & 1.00\% \\ 
\hline
\end{tabular}
\caption{FerrLive and FerrFake for submitted system in LivDet 2015.}
\label{tab:x}
\end{center}
\end{table} 

\section{Quality analysis: A Lesson Learned from the LivDet Experience}
\label{sec:lessonLearned}

%The LivDet organization and implementation require a big effort in terms of time and resources. It includes volunteer scheduling, acquisition of their fingerprints, and quality checks of spoofs and images. 

This section explores the impact of spoof quality on  algorithm performance.

Over a time span of ten years, the researchers in the respective groups have searched for materials suitable for creating fingerprint spoofs by experimenting with a very large number of materials and related variants.  This process is performed independent of the feature extraction algorithms and the classification methods adopted.  In general, it is difficult to create a spoof fingerprint that gives good quality fingerprint images on a consistent basis.  The quality analysis here reflects this experience.

With regard to the observation above, Tables \ref{tab:qualityDegree}, \ref{tab:easynessDegree} report a summary of our research. They include a certain number of materials used as cast and/or mold. Table \ref{tab:qualityDegree} summarizes the subjective evaluation made by our stuff over ten years experience, and more than 50000 spoofs realized, of the visual quality of the obtained fake fingerprints when combining the mold material (columns) and the cast material (rows). The visual quality assessment has been made up on the basis of the analysis at the microscope of the spoofs and also the quality of images acquired by the capture devices. We experienced that certain materials are less suitable than others. Several of them are not able to replicate the ridges and valleys flow without adding evident artifacts as bubbles and altering the ridges edges.

\begin{table*}[p]
\begin{center}
\begin{tabular}[t]{ | l || c | c | c | c |}\hline
\textbf{Mold/Material}	& \textbf{Foil} & \textbf{RTV}	& \textbf{Modasil}	& \textbf{Plasticine}\\ \hline \hline
Alginat		&		&	5	&	5	&		\\ \hline
RTV			&	2	&	2	&	4	&		\\ \hline
Ecoflex		&	4	&	1	&	3	&		\\ \hline
Gelatine	&	2	&	1	&	3	&		\\ \hline
Latex		&	2	&	1	&	3	&		\\ \hline
Modasil		&	2	&	2	&	4	&		\\ \hline
Transparent	&	3	&		&		&	5	\\
Silicone	&	&	&	&	\\ \hline
White		&	2	&		&		&		\\ 
Silicone	&	&	&	&	\\ \hline
Wood glue	&	2	&	1	&	3	&		\\ \hline
SILIGUM		&	2	&		&		&	5	\\
\hline
\end{tabular}
\caption{Quality degree in the spoof fabrication process: 1-Very high, 2-High, 3-Medium, 4-Low, 5-Very Low.}
\label{tab:qualityDegree}
\end{center}
\end{table*}

\begin{table*}[p]
\begin{center}
\begin{tabular}[t]{ | l || c | c | c | c |}\hline
\textbf{Mold/Material}	& \textbf{Foil} & \textbf{RTV}	& \textbf{Modasil}	& \textbf{Plasticine}\\ \hline \hline
Alginat		&		&	5	&	5	&		\\ \hline
RTV			&	1	&	4	&	3	&		\\ \hline
Ecoflex		&	1	&	4	&	4	&		\\ \hline
Gelatine	&	1	&	4	&	3	&		\\ \hline
Latex		&	1	&	4	&	3	&		\\ \hline
Modasil		&	1	&	4	&	3	&		\\ \hline
Transparent	&	1	&		&		&	3	\\
Silicone	&	&	&	&	\\ \hline
White		&	1	&		&		&		\\ 
Silicone	&	&	&	&	\\ \hline
Wood glue	&	1	&	4	&	4	&		\\ \hline
SILIGUM		&	1	&		&		&	3	\\
\hline
\end{tabular}
\caption{Easiness degree in the fake fabrication process: 1-Very High, 2-High, 3-Medium, 4-Low, 5-Very Low.}
\label{tab:easynessDegree}
\end{center}
\end{table*}

Table \ref{tab:easynessDegree} shows the subjective evaluation on the easiness of obtaining a good spoof from the combination of the same materials of Table \ref{tab:qualityDegree}. This evaluation depends on the solidification time of the adopted material, the level of difficulty in separating mold and cast without destroying one of them or both, the natural dryness or wetness level of the related spoof.

Tables \ref{tab:qualityDegree}, \ref{tab:easynessDegree} show that, from a practical viewpoint, many materials are difficult to manage when fabricating a fake finger. In many cases, the materials with this property also exhibit a low subjective quality level.

Therefore, thanks to this lesson, the LivDet competition challenge participants with images coming from the spoofs obtained with the best and most ``potentially dangerous'' materials. The materials choice is made on the basis of the best trade off between the criteria pointed out in Tables \ref{tab:qualityDegree}, \ref{tab:easynessDegree} and the objective quality values output by quality assessment algorithms such as NFIQ.

\begin{figure}[t]
\begin{center}
%  \includegraphics[width=1\linewidth]{spoof13.eps}
%\subfloat{\includegraphics[width=0.45\linewidth]{ferrFakeDerm.eps}}
\subfloat{\includegraphics[width=0.45\linewidth]{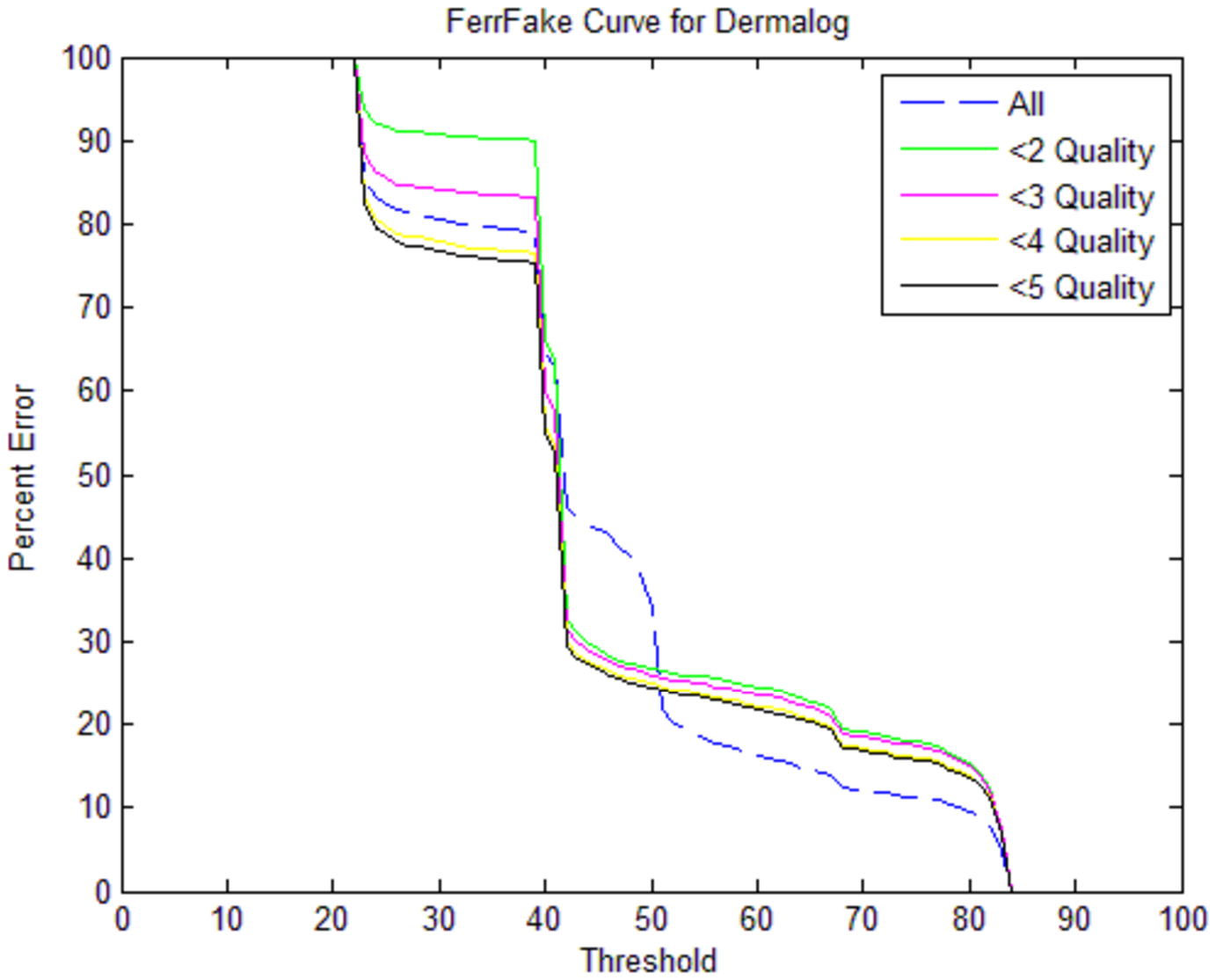}}
\qquad
%\subfloat{\includegraphics[width=0.45\linewidth]{ferrLiveDerm.eps}}
\subfloat{\includegraphics[width=0.45\linewidth]{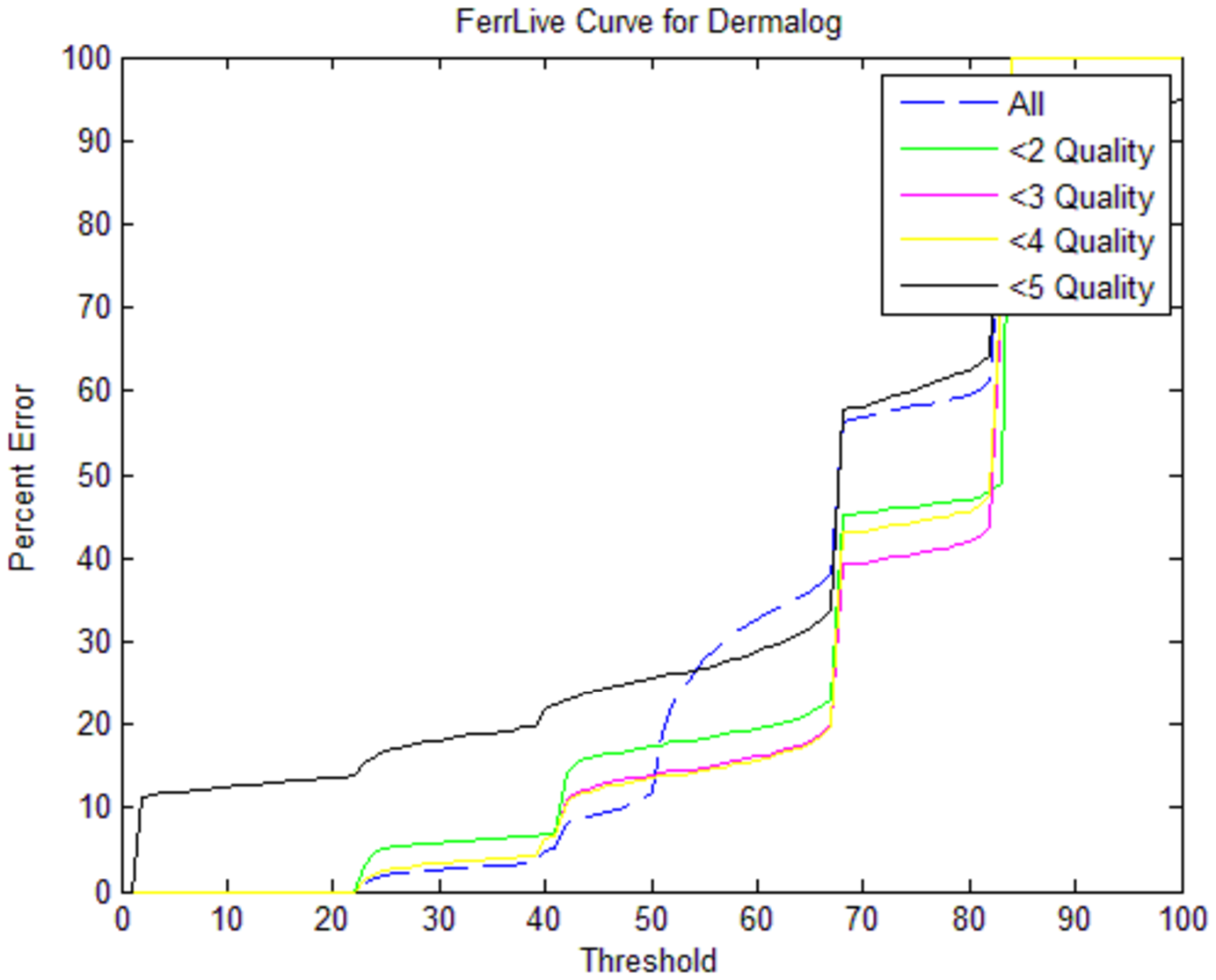}}
\end{center}
\caption{Error rates by quality level.}
\label{fig:ferrDerm}       % Give a unique label
\end{figure}

What reported has been confirmed along the four LivDet editions. In particular the FerrFake and FerrLive rates for each differing quality levels support the idea that the images quality level is correlated with the error rate decrease. The error rates for each range of quality levels for Dermalog in LivDet 2011 Fingerprint Part 1: Algorithms is shown in figure \ref{fig:ferrDerm}, as an example. The graphs showcase from images of only quality level 1 up to all quality levels being shown. As lower quality spoof images were added, ferrfake generally decreased. For all images which included the worst quality images, the error rates were less consistent likely due to the variability in low quality spoofs.  

%Examining the data from LivDet 2013 Part 1: Algorithms, it can be shown that the Crossmatch dataset followed along the lines of the data in LivDet 2011 with high percentages of the data being in the top two quality levels.  The swipe dataset had many images that were read as being of lower quality which could be seen in the data itself because of the difficulty in collecting spoof data on the swipe device. Figures \ref{fig:spoofPerc} and \ref{fig:imagePerc} shows the percentage of images by quality level in LivDet 2013.

\begin{figure}[t]
\begin{center}
  \includegraphics[width=1\linewidth]{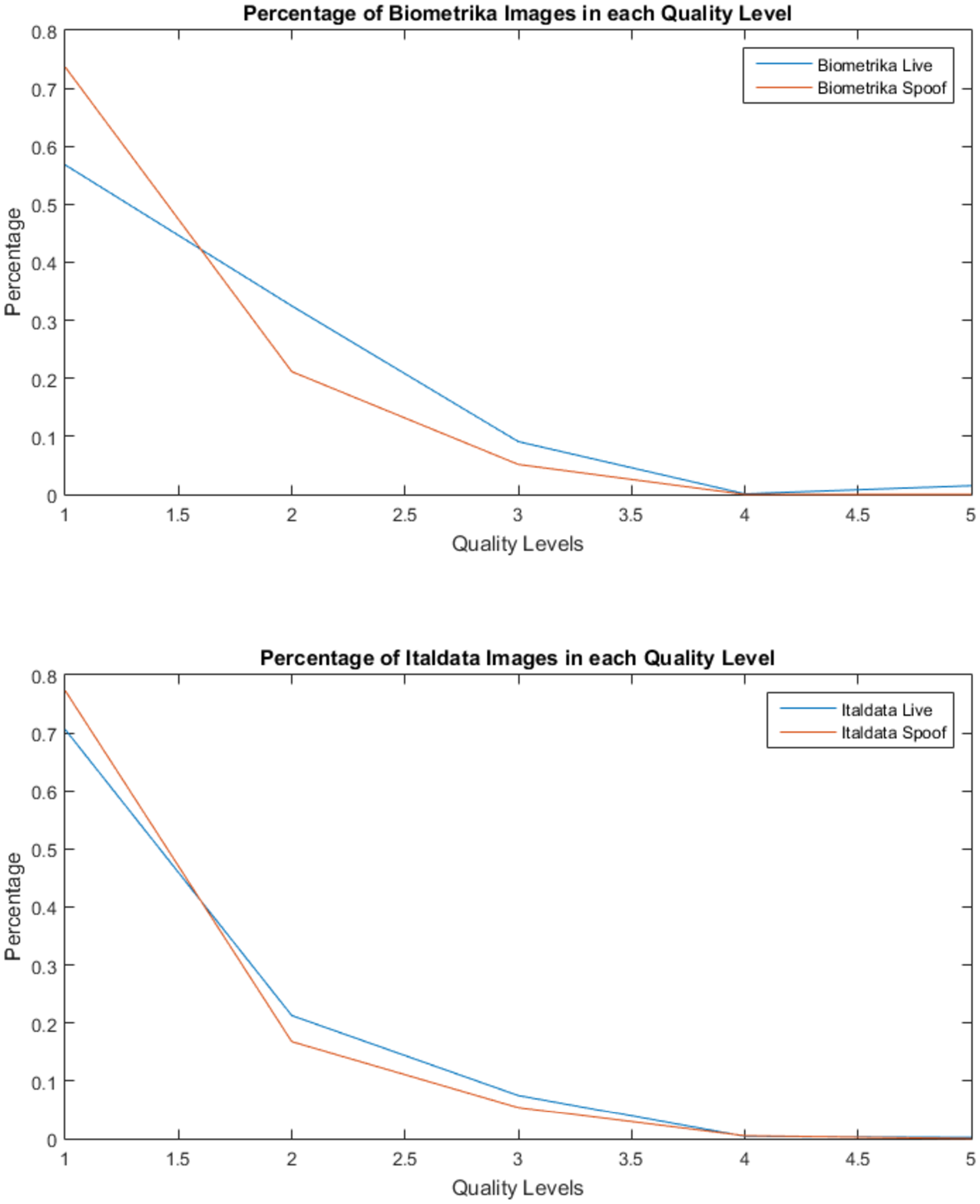}
\end{center}
\caption{Percentages of Images per Quality Level for LivDet 2011.}
\label{fig:imagePerc}       % Give a unique label
\end{figure}

\begin{figure}[t]
\begin{center}
  \includegraphics[width=1\linewidth]{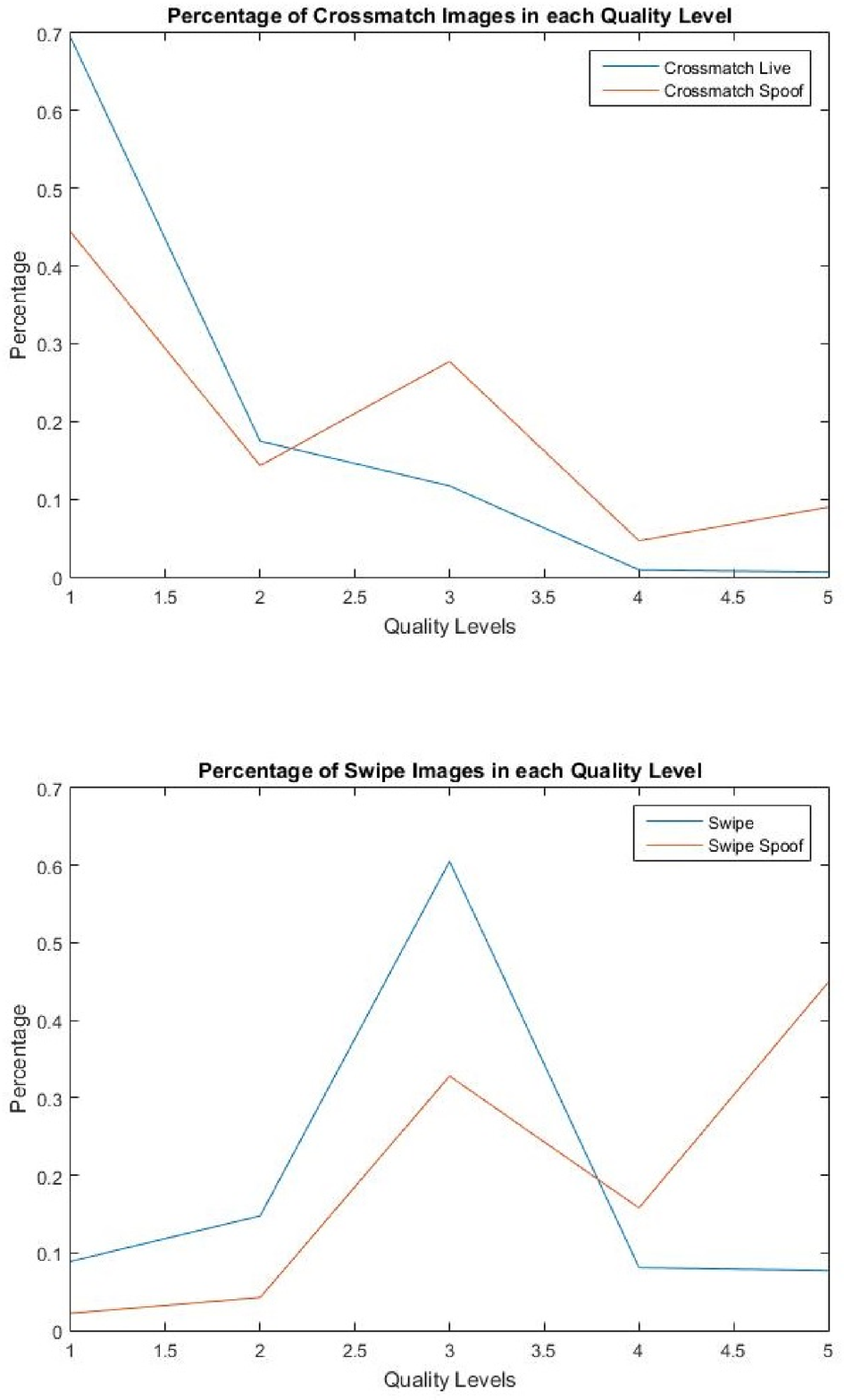}
\end{center}
\caption{Percentages of images per Quality Level for LivDet 2013.}
\label{fig:spoofPerc}       % Give a unique label
\end{figure}

\begin{figure}[t]
\begin{center}
  \includegraphics[width=1\linewidth]{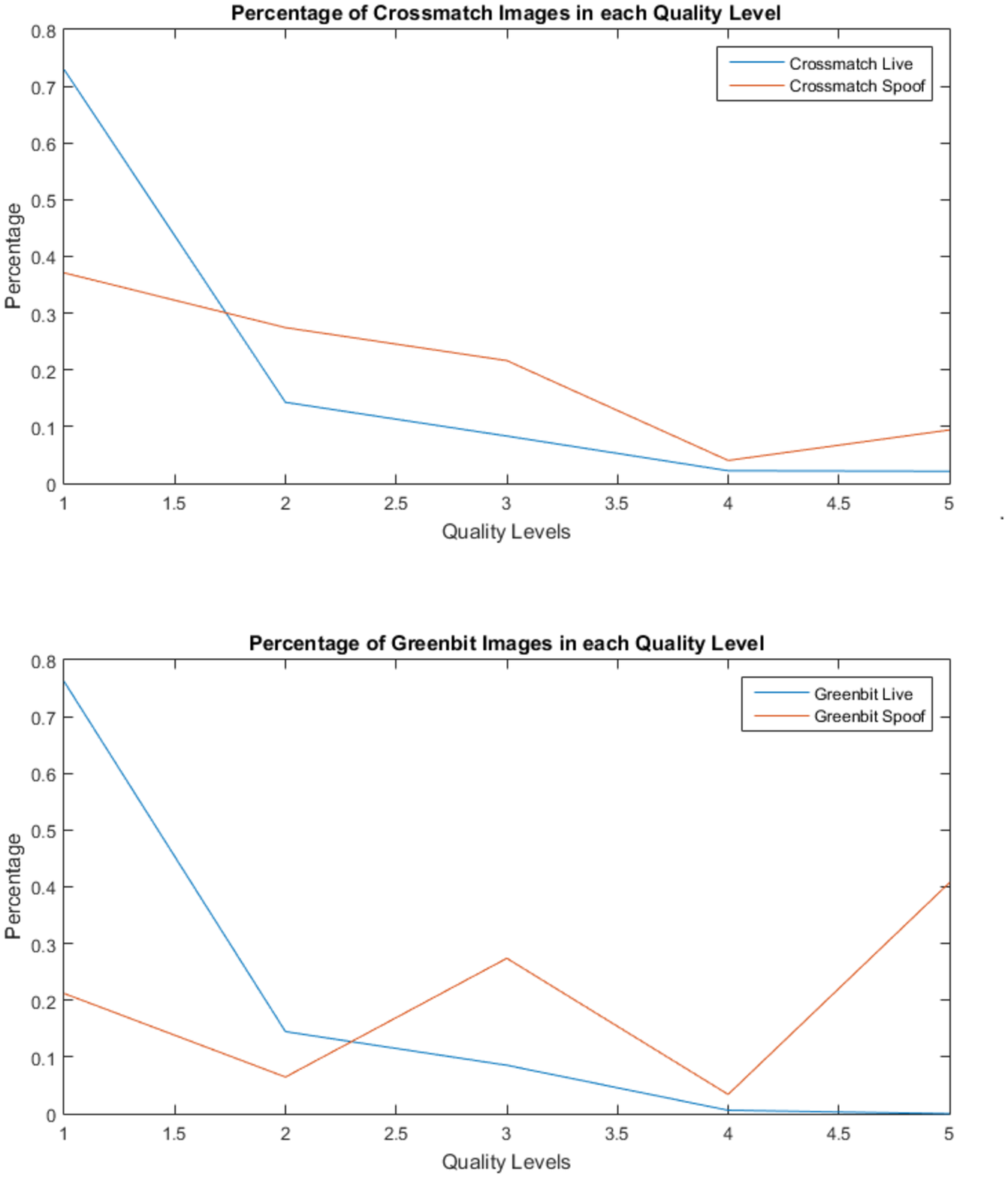}
\end{center}
\caption{Percentages of images per Quality Level for LivDet 2015.}
\label{fig:quality2015}       % Give a unique label
\end{figure}

The percentage of images at each quality level for two representative datasets for LivDet 2011, 2013, and 2015, respectively, are given in Figures \ref{fig:imagePerc}, \ref{fig:spoofPerc}, and \ref{fig:quality2015}.   The Crossmatch dataset had high percentages of the data being in the top two quality levels in both LivDet 2011 and 2013. The swipe dataset had many images that were read as being of lower quality which could be seen in the data itself because of the difficulty in collecting spoof data on the swipe device. 

\section{Conclusions}
\label{sec:conclusion}

Since its first edition in 2009, the Fingerprint Liveness Detection Competition was aimed to allow research centres and companies a fair and independent assessment of their anti-spoofing algorithms and systems.

We have seen over time an increasing interest for this event, and the general recognition for the enormous amount of data made publicly available. The number of citations that LivDet competitions have collected is one of the tangible signs of such interest (about 100 citations according to Google Scholar) and further demonstrates the benefits that the scientific community has received from LivDet events.

The competition results show that liveness detection algorithms and systems strongly improved their performance: from about  70\% classification accuracy achieved in LivDet 2011, to  90\% classification accuracy in LivDet 2015. This result, obtained under very difficult conditions like the ones of the consensual methodology of fingerprints replication, is comparable with that obtained in LivDet 2013 (first two data sets), where the algorithms performance was tested under the easier task of fingerprints replication from latent marks. Moreover, the two challenges characterizing the last edition, namely, the presence of 1000 dpi capture device and the evaluation against ``unknown" spoofing materials, further contributed to show the great improvement that researchers achieved on these issues: submitted algorithms performed very well on both 500 and 1000 dpi capture devices, and some of them also exhibited a good robustness degree against never-seen-before attacks.
Results reported on fusion also shows that the liveness detection could further benefit from the combination of multiple features and approaches. A specific section on algorithms and systems fusion might be explicitly added to a future LivDet edition.

There is a dark side of the Moon, of course.
It is evident that, despite the remarkable results reported in this paper, there is a clear need of further improvements. Current performance for most submissions are not yet good enough for embedding a liveness detection algorithm into fingerprint verification system where the error rate is still too high for many real applications.
In the authors' opinion, discovering and explaining benefits and limitations of the currently used features is still an issue whose solution should be encouraged, because only the full understanding of the physical process which leads to the finger's replica and what features extraction process exactly does will shed light on the characteristics most useful for classification. We are aware that this is a challenging task, and many years could pass before seeing concrete results. However, we believe this could be the next challenge for a future edition of LivDet, the Fingerprint Liveness Detection Competition.

\section*{Acknowledgements}
The first and second author had equal contributions to the research.
This work has been supported by the Center for Identification Technology Research and the National Science Foundation under Grant No. 1068055, and by the project ``Computational quantum structures at the service of pattern recognition: modeling uncertainty" [CRP-59872] funded by Regione Autonoma della Sardegna, L.R. 7/2007, Bando 2012.

\section*{References}

%\end{acknowledgements}



\end{document}